\documentclass[journal]{IEEEtran}
\usepackage{amsmath,amsfonts}
\usepackage{array}
\usepackage{algorithmic}
\usepackage[ruled,lined,linesnumbered,ruled]{algorithm2e}  
\usepackage{multirow}
\usepackage{graphicx} 
\usepackage{float} 
\usepackage[caption=false,font=footnotesize,labelfont=rm,textfont=rm]{subfig}
\usepackage{CJKutf8}
\usepackage{textcomp}
\usepackage{stfloats}
\usepackage{url}
\usepackage{verbatim}
\usepackage{graphicx}
\usepackage{cite}
\begin{document}
\begin{CJK}{UTF8}{gbsn}
\title{\textbf{Fuzzy Recurrent Stochastic Configuration Networks for Industrial Data Analytics}}
\author{
  Dianhui Wang 
  \thanks{\textit{\underline{Corresponding author}}: 
\textbf{dh.wang@deepscn.com}}\\
  State Key Laboratory of Synthetical Automation for Process Industries \\
  Northeastern University, Shenyang 110819, China \\
 Research Center for Stochastic Configuration Machines\\
  China University of Mining and Technology, Xuzhou 221116, China\\
  Gang Dang \\
  State Key Laboratory of Synthetical Automation for Process Industries \\
  Northeastern University, Shenyang 110819, China\\
}
\maketitle
\newtheorem{remark}{\bf Remark}

\begin{abstract}
This paper presents a novel neuro-fuzzy model, termed fuzzy recurrent stochastic configuration networks (F-RSCNs), for industrial data analytics. Unlike the original recurrent stochastic configuration network (RSCN), the proposed F-RSCN is constructed by multiple sub-reservoirs, and each sub-reservoir is associated with a Takagi–Sugeno–Kang (TSK) fuzzy rule. Through this hybrid framework, first, the interpretability of the model is enhanced by incorporating fuzzy reasoning to embed the prior knowledge into the network. Then, the parameters of the neuro-fuzzy model are determined by the recurrent stochastic configuration (RSC) algorithm. This scheme not only ensures the universal approximation property and fast learning speed of the built model but also overcomes uncertain problems, such as unknown dynamic orders, arbitrary structure determination, and the sensitivity of learning parameters in modelling nonlinear dynamics. Finally, an online update of the output weights is performed using the projection algorithm, and the convergence analysis of the learning parameters is given. By integrating TSK fuzzy inference systems into RSCNs, F-RSCNs have strong fuzzy inference capability and can achieve sound performance for both learning and generalization. Comprehensive experiments show that the proposed F-RSCNs outperform other classical neuro-fuzzy and non-fuzzy models, demonstrating great potential for modelling complex industrial systems.
\end{abstract}


\section{Introduction}
 \IEEEPARstart{R}{ecently}, complex patterns and dynamics of industrial processes such as electricity, chemical industry, metallurgy, and machinery manufacturing pose significant challenges in developing precise physical models. Neuro-fuzzy modelling techniques offer an alternative approach to approximate intricate nonlinear systems \cite{ref1, ref2, ref3, ref4}. They integrate the strengths of both neural networks (NNs) and fuzzy inference systems, which have strong nonlinear processing capability, logical reasoning capability, and interpretability \cite{ref5, ref6, ref7}. However, traditional neuro-fuzzy models cannot automatically update their rules or parameters based on changing data patterns, indicating a lack of adaptive learning capability \cite{ref8, ref9, ref10}. Furthermore, they typically require multiple iterations to establish a comprehensive set of rules, resulting in a heavy computational burden \cite{ref11, ref12, ref121}. In industrial environments, computing resources are often limited, particularly on edge computing devices \cite{ref13}. Therefore, it is essential to develop lightweight learning models and interpretable artificial intelligence (AI) for practical industrial applications.

Industrial systems encompass a variety of dynamic processes, with input orders at different stages evolving over time. Temporal data analysis strategies can effectively capture the time-dependent trends and patterns within these systems, facilitating the modelling of dynamic processes characterized by unknown or uncertain orders \cite{ref141, ref14, ref15, ref16, ref17}. Among them, recurrent neural networks (RNNs) have feedback connections between neurons, which can handle the uncertainty caused by the selected input variables that are fed into a learner model. Long short-term memory (LSTM) networks introduce more complex unit structures and gating mechanisms that enhance control over the flow and retention of information. Unfortunately, training RNNs and LSTM networks by the back-propagation (BP) algorithm suffers from the sensitivity of learning rate, slow convergence, and local minima. As the volume and complexity of data increase, they face scalability issues, leading to significantly greater training time and resource requirements. This limitation restricts their applicability in large-scale industrial applications. With such a background, randomized learner models have been developed over the past years. Jaeger \cite{ref20} demonstrated that echo state networks (ESNs) could utilize large-scale sparsely connected reservoirs to capture and process temporal information, effectively overcoming the drawbacks of the BP-based method and having been widely applied for modelling complex dynamics \cite{ref21, ref211, ref212, ref213, ref214}. However, ESNs adopt the fixed scope setting of the random parameters, which cannot be adaptively adjusted based on the input data. Further optimization processes are required to determine the optimal network topology and parameters. In addition, all the aforementioned models present several challenges, including arbitrary structure determination, a lack of theoretical guarantees that the resulting model can effectively approximate nonlinear dynamic systems, and limited interpretability. These uncertainties in the training processes can result in unstable predictions, potentially undermining the reliability of industrial systems.

To ensure the universal approximation performance of the randomized NNs for any nonlinear maps, Wang and Li introduced an advanced randomized learner model, termed stochastic configuration networks (SCNs) \cite{ref22}. The unique supervisory mechanism of SCNs not only aids in generating appropriate network structures but also significantly reducing the burden of parameter tuning. Some noteworthy studies of the SCN framework include the deep SCN with multiple hidden layers for better modelling performance, 2D-SCN for image data processing, robust SCN for resolving uncertain noisy data regression problems, and fuzzy SCN for improved interpretability have been reported in \cite{ref23, ref24, ref25, ref26}. Although SCNs and their variants have shown promise in resolving industrial data modelling tasks, they face challenges in modelling nonlinear systems with unknown dynamic orders. In \cite{ref27}, we proposed the recurrent stochastic configuration networks (RSCNs), where the system's dynamic order information is not requested and a special feedback matrix is defined to ensure the universal approximation property for offline modelling. RSCNs offer advantages such as strong learning capability, computational simplicity, and a good potential to implement lightweight recurrent models. However, they have limitations in dealing with complex fuzzy systems and lack interpretability, which impact the decision-making process of the system. This paper presents a fuzzy version of RSCNs (F-RSCNs) to enhance the modelling capability of RSCNs for uncertain and fuzzy nonlinear industry systems, while also addressing the drawbacks of traditional neuro-fuzzy models in terms of modelling efficiency and structure design. The proposed approach has the following advantages.
\begin{itemize}
  \item [1)] 
  F-RSCNs can guarantee the universal approximation property of the model in offline modelling, as well as the convergence and stability of the network parameters during dynamic learning.       
  \item [2)]
  The nonlinear parameters can be generated adaptively through stochastic configuration rather than the BP algorithm, significantly promoting the learning speed. 
  \item [3)]
  For temporal data modelling, it is sufficient to understand the delay characteristics of the system input, without requiring knowledge of all the dynamic orders.
  \item [4)]
  F-RSCNs combine the strengths of both RSCNs and fuzzy inference systems, exhibiting strong predictive capabilities for the real industrial process data.
\end{itemize}  
 
The remainder of this paper is organized as follows. Section II reviews the relevant background of fuzzy inference systems, SCNs, and ESNs. Section III details the proposed F-RSCNs with both algorithm description and theoretical analysis. Section IV reports our experimental results with comparisons and discussions. Finally, Section V concludes this paper and provides further research direction.

\section{Related work}
This section introduces some related work, including the well-known Takagi-Sugeno-Kang (TSK) fuzzy inference systems, stochastic configuration networks, and echo state networks.
\subsection{TSK fuzzy inference systems}
TSK fuzzy inference system is a classical fuzzy framework \cite{ref28}, which is based on the fuzzy logic theory and is usually used for handling fuzzy rule-based modelling tasks. Unlike the traditional Mamdani fuzzy inference systems, TSK-based approaches represent the outputs by combining linear functions rather than fuzzy sets. Moreover, they can easily explain some fuzzy and qualitative knowledge, making them suitable for applications in nonlinear system identification, process control, and modelling \cite{ref5, ref6, ref15, ref16, ref17, ref26, ref29}.

The TSK fuzzy inference system is composed of antecedents and consequent components. The antecedent part is responsible for matching the fuzzy rules with the input data, and evaluating the degree of applicability for each rule. The consequent part is used to aggregate the output of fuzzy rules. The system output is the weighted sum of the results for each fuzzy rule, where the weights are the applicable values calculated in the antecedent part. Specifically, a typical TSK fuzzy rule can be described as
\begin{equation}
\label{eq1}
\begin{array}{l}
{\rm{Rul}}{{\rm{e}}^i}:{\rm{If}}{\kern 1pt} {\kern 1pt} {\kern 1pt} {u_1}{\kern 1pt} {\kern 1pt} {\kern 1pt} {\rm{is}}{\kern 1pt} {\kern 1pt} {\kern 1pt} {A_i},{\kern 1pt} {\kern 1pt} {\kern 1pt} {u_2}{\kern 1pt} {\kern 1pt} {\kern 1pt} {\rm{is}}{\kern 1pt} {\kern 1pt} {\kern 1pt} {A_i},{\kern 1pt} {\kern 1pt}  \ldots ,{\kern 1pt} {\kern 1pt} {\kern 1pt} {\rm{and}}{\kern 1pt} {\kern 1pt} {\kern 1pt} {u_K}{\kern 1pt} {\kern 1pt} {\kern 1pt} {\rm{is}}{\kern 1pt} {\kern 1pt} {\kern 1pt} {A_i},\\
{\kern 1pt} {\rm{Then}}:{y_i} = {\zeta ^i}\left( {{u_1},{u_2}, \ldots ,{u_K}} \right),{\kern 1pt} {\kern 1pt} {\kern 1pt} i = 1,2, \ldots ,Q,
\end{array}
\end{equation}
where $\mathbf{u}=\left[ {{u}_{1}},{{u}_{2}},\ldots ,{{u}_{K}} \right]^{\top }\in {{\mathbb{R}}^{K}}$ is the input signal, $K$ is the dimension of the input, ${{A}_{i}}$ denotes the fuzzy variable, $Q$ is the number of fuzzy rules, ${{y}_{i}}$ is the output of the $i\text{-th}$ fuzzy rule, and ${{\zeta }^{i}}$ is a polynomial of the input variables. The system output is a linear combination of ${{\zeta }^{1}},{{\zeta }^{2}},\ldots ,{{\zeta }^{Q}}$, and the weights are evaluated by membership functions. Gaussian functions provide a general scheme to express the membership degree of each input variable to the fuzzy set: 
\begin{equation}
\label{eq2}
{{\mu }_{{{A}_{i}}}}\left( {{u}_{k}} \right)=\exp \left( -{{\left( \frac{{{u}_{k}}-c_{k}^{i}}{\sigma _{k}^{i}} \right)}^{2}} \right),k=1,2,\ldots ,K
\end{equation}
where $c_{k}^{i}$ and $\sigma _{k}^{i}$ are the center and width of the $i\text{-th}$ membership function for the $k\text{-th}$ input. The fire strength of the $i\text{-th}$ fuzzy rule is given by
\begin{equation}
\label{eq3}
{{\psi }_{i}}=\prod\limits_{k=1}^{K}{{{\mu }_{{{A}_{i}}}}\left( {{u}_{k}} \right)},i=1,2,\ldots ,Q.
\end{equation}
Then, the output of the TSK fuzzy inference system can be calculated by
\begin{equation}
\label{eq4}
y=\frac{\sum\limits_{i=1}^{Q}{{{\psi }_{i}}{{y}_{i}}}}{\sum\limits_{i=1}^{Q}{{{\psi }_{i}}}}=\frac{\sum\limits_{i=1}^{Q}{\prod\limits_{k=1}^{K}{{{\mu }_{{{A}_{i}}}}\left( {{u}_{k}} \right)}{{\zeta }^{i}}\left( {{u}_{1}},{{u}_{2}},\ldots ,{{u}_{K}} \right)}}{\sum\limits_{i=1}^{Q}{\prod\limits_{k=1}^{K}{{{\mu }_{{{A}_{i}}}}\left( {{u}_{k}} \right)}}}.
\end{equation}

\subsection{Stochastic configuration networks}
Stochastic configuration networks (SCNs) \cite{ref22} are a class of randomized learner models, where the random weights and biases are incrementally generated in the light of a supervisory mechanism, ensuring the universal approximation property of the built model. Additionally, the stochastic configuration (SC) algorithm can be easily implemented, and solve the problems of network parameter selection and arbitrary structure determination. Due to their advantages of high learning efficiency, low human intervention, and strong approximation ability, SCNs have attracted considerable interest in modelling uncertain nonlinear systems, and many promising results have been reported \cite{ref25, ref26, ref301}. For more details about SCNs, one can refer to \cite{ref22}.

Given an objective function $f:{{\mathbb{R}}^{K}}\to {{\mathbb{R}}^{L}}$, ${{n}_{\max }}$ groups training samples $\left\{ \mathbf{u}\left( n \right),\mathbf{t}\left( n \right) \right\}$, $\mathbf{u}\left( n \right)=\left[ {{u}_{1}}\left( n \right),\ldots ,{{u}_{K}}\left( n \right) \right]^{\top }$, $\mathbf{t}\left( n \right)=\left[ {{t}_{1}}\left( n \right),\ldots ,{{t}_{L}}\left( n \right) \right]^{\top }$, $\mathbf{U}=\left[ \mathbf{u}\left( 1 \right),\ldots ,\mathbf{u}\left( {{n}_{\max }} \right) \right]\in {{\mathbb{R}}^{K\times {{n}_{\max }}}}$ is the input data, and the corresponding output is $\mathbf{T}=\left[ \mathbf{t}\left( 1 \right),\ldots ,\mathbf{t}\left( {{n}_{\max }} \right) \right]\in {{\mathbb{R}}^{L\times {{n}_{\max }}}}$, $K$ and $L$ are the dimensions of the input and output, respectively. Suppose we have built a single-layer feedforward network with $N-1$ hidden nodes,
\begin{small}
\begin{equation}
\label{eq5}
{{f}_{N-1}}=\sum\limits_{j=1}^{N-1}{{{\mathbf{\beta }}_{j}}{{g}_{j}}\left( \mathbf{w}_{j}^{\top }\mathbf{U}+{{\mathbf{b}}_{j}} \right)}\left( {{f}_{0}}=0,N=1,2,3... \right),
\end{equation}
\end{small}
where ${{\mathbf{w}}_{j}}$, ${{\mathbf{b}}_{j}}$, and ${{\mathbf{\beta }}_{j}}$ are the input weight, bias, and output weight of the $j\text{-th}$ hidden node, ${{g}}$ is the activation function. The residual error between the current model output and the expected output ${{f}_{\text{exp}}}$ is defined as
\begin{equation}
\label{eq6}
{{e}_{N-1}}={{f}_{\text{exp}}}-{{f}_{N-1}}=\left[ {{e}_{N-1,1}},{{e}_{N-1,2}},...{{e}_{N-1,L}} \right].
\end{equation}
If ${{e}_{N-1}}$ fails to reach the preset error tolerance, it is necessary to configure a new random basis function ${{g}_{N}}$ based on the supervisory mechanism. Then, we evaluate the output weight ${{\mathbf{\beta }}_{N}}$ and update the model ${{f}_{N}}={{f}_{N-1}}+{{\mathbf{\beta }}_{N}}{{g}_{N}}$. Repeat the configuration until satisfy the terminal conditions. 

In \cite{ref22}, three SC algorithms are presented to learn the network parameters, and numerous simulation results have indicated that the SC-III algorithm outperforms others with better learning ability and sound generalization. Therefore, the SC-based networks mentioned in this paper are all trained using the SC-III algorithm. The basic theoretical description of SCNs is shown in Theorem 1.\\ 

\textbf{Theorem 1 \cite{ref22}.} Let span(Γ) be dense in ${{L}_{2}}$ space and $\forall g\in \Gamma ,0<\left\| g \right\|<{{b}_{g}}$, ${{b}_{g}}\in {{\mathbb{R}}^{+}}$. Given $0<r<1$ and a non-negative real number sequence $\left\{ {{\mu }_{N}} \right\}$ , satisfy $\underset{N\to \infty }{\mathop{\lim }}\,{{\mu }_{N}}=0$ and ${{\mu }_{N}}\le (1-r)$. For $N\text{=}1,2,...$, define
\begin{small}
\begin{equation}
\label{eq7}
{{\delta }_{N}}=\sum\limits_{q=1}^{L}{\delta _{N,q}^{{}}},\delta _{N,q}^{{}}=(1-r-{{\mu }_{N}}){{\left\| {{e}_{N-1,q}} \right\|}^{2}},q=1,2,\ldots ,L.
\end{equation}
\end{small}
If the generated random basis function ${{g}_{N}}$ satisfies the following inequality constraint 
\begin{equation}
\label{eq8}
{{\left\langle {{e}_{N-1,q}},{{g}_{N}} \right\rangle }^{2}}\ge b_{g}^{2}\delta _{N,q}^{{}},q=1,2,\ldots ,L,
\end{equation}
and the output weight is determined by the global least square method 
\begin{equation}
\label{eq9}
\left[ \mathbf{\beta }_{1}^{*},\mathbf{\beta }_{2}^{*},...,\mathbf{\beta }_{N}^{*} \right]=\underset{\mathbf{\beta }}{\mathop{\arg \min }}\,\left\| {{f}_{\exp }}-\sum\limits_{j=1}^{N}{{{\mathbf{\beta }}_{j}}{{g}_{j}}} \right\|,
\end{equation}
we have $\underset{N\to \infty }{\mathop{\lim }}\,\left\| {{f}_{\text{exp}}}-{{f}_{N}} \right\|=0$.

\subsection{Echo state networks }
ESNs randomly assign the input and reservoir weights in a fixed uniform distribution, with only the output weights being calculated. The readout can be obtained using a linear combination of the reservoir states. ESNs simplify the training process of RNNs, solving the problems of complex parameter updates, excessive computational burden, and local minima. Moreover, ESNs have the echo state property \cite{ref30}, that is, the reservoir state $\mathbf{x}\left( n \right)$ is the echo of the input and $\mathbf{x}\left( n \right)$ should asymptotically depend on the driving input signal. This distinctive property makes them highly promising for handling temporal data.

Consider an ESN model
\begin{equation}
\label{eq10}
\mathbf{x}(n)=g({{\mathbf{W}}_{\text{in}}}\mathbf{u}(n)+{{\mathbf{W}}_{\operatorname{r}}}\mathbf{x}(n-1)+\mathbf{b}),
\end{equation}
\begin{equation}
\label{eq11}
\mathbf{y}(n)={{\mathbf{W}}_{\text{out}}}\left( \mathbf{x}(n),\mathbf{u}(n) \right),
\end{equation}
where $\mathbf{u}(n)\in {{\mathbb{R}}^{K}}$ is the input signal at time step $n$; $\mathbf{x}(n)\in {{\mathbb{R}}^{N}}$ is the internal state of the reservoir; ${{\mathbf{W}}_{\text{in}}}\in {{\mathbb{R}}^{N\times K}},{{\mathbf{W}}_{\operatorname{r}}}\in {{\mathbb{R}}^{N\times N}}$ represent the input and reservoir weights, respectively; $\mathbf{b}$ is the bias; ${{\bf{W}}_{{\rm{out}}}} \in {^{L \times \left( {N + K} \right)}}$ is the output weight; and $g$ is the activation function selected from the hyperbolic tangent function or sigmoid function.  $\mathbf{x}(0)$ is started with a zero matrix, and a few samples are usually washed out for minimizing the influence of the initial zero states. ${{\mathbf{W}}_{\text{in}}},{{\mathbf{W}}_{\text{r}}}$, and $\mathbf{b}$ are generated from the uniform distribution $\left[ -\lambda ,\lambda  \right].$ Notably, the value of $\lambda $ has a significant impact on model performance. The original ESNs adopt a fixed $\lambda $, which may lead to poor performance. Scholars have focused on optimizing the weight scope, and some promising results have been reported in \cite{ref31, ref32}. However, the optimization process inevitably increases the complexity of the algorithm. Therefore, selecting a data-dependent and adjustable $\lambda $ is crucial.

Define $\mathbf{X}\text{=}\left[ \left( \mathbf{x}(1),\mathbf{u}(1) \right),\ldots ,\left( \mathbf{x}({{n}_{\max }}),\mathbf{u}({{n}_{\max }}) \right) \right]$, where ${{n}_{\max }}$ is the number of training samples and the output is
\begin{equation}
\label{eq12}
\mathbf{Y}=\left[ \mathbf{y}\left( 1 \right),\mathbf{y}\left( 2 \right),...\mathbf{y}\left( {{n}_{max}} \right) \right]=\mathbf{W}_{\operatorname{out}}^{{}}\mathbf{X}.
\end{equation}
The output weight ${{\mathbf{W}}_{\rm{out}}}$ can be calculated by the least squares method 
\begin{equation}
\label{eq13}
\underset{\mathbf{W}_{\operatorname{out}}^{{}}}{\mathop{\min }}\,\left\| \mathbf{W}_{\operatorname{out}}^{{}}\mathbf{X}-\mathbf{T} \right\|^{2},
\end{equation}
\begin{equation}
\label{eq14}
\mathbf{W}_{\operatorname{out}}^{\top }={{\left( \mathbf{X}{{\mathbf{X}}^{\top }} \right)}^{-1}}\mathbf{X}{{\mathbf{T}}^{\top }},
\end{equation}
where $\mathbf{T}=\left[ \mathbf{t}\left( 1 \right),...\mathbf{t}\left( {{n}_{max}} \right) \right]$ is the desired output. 

\section{Fuzzy recurrent stochastic configuration networks}
\begin{figure}[ht]
\centering
\includegraphics[width=3.5in]{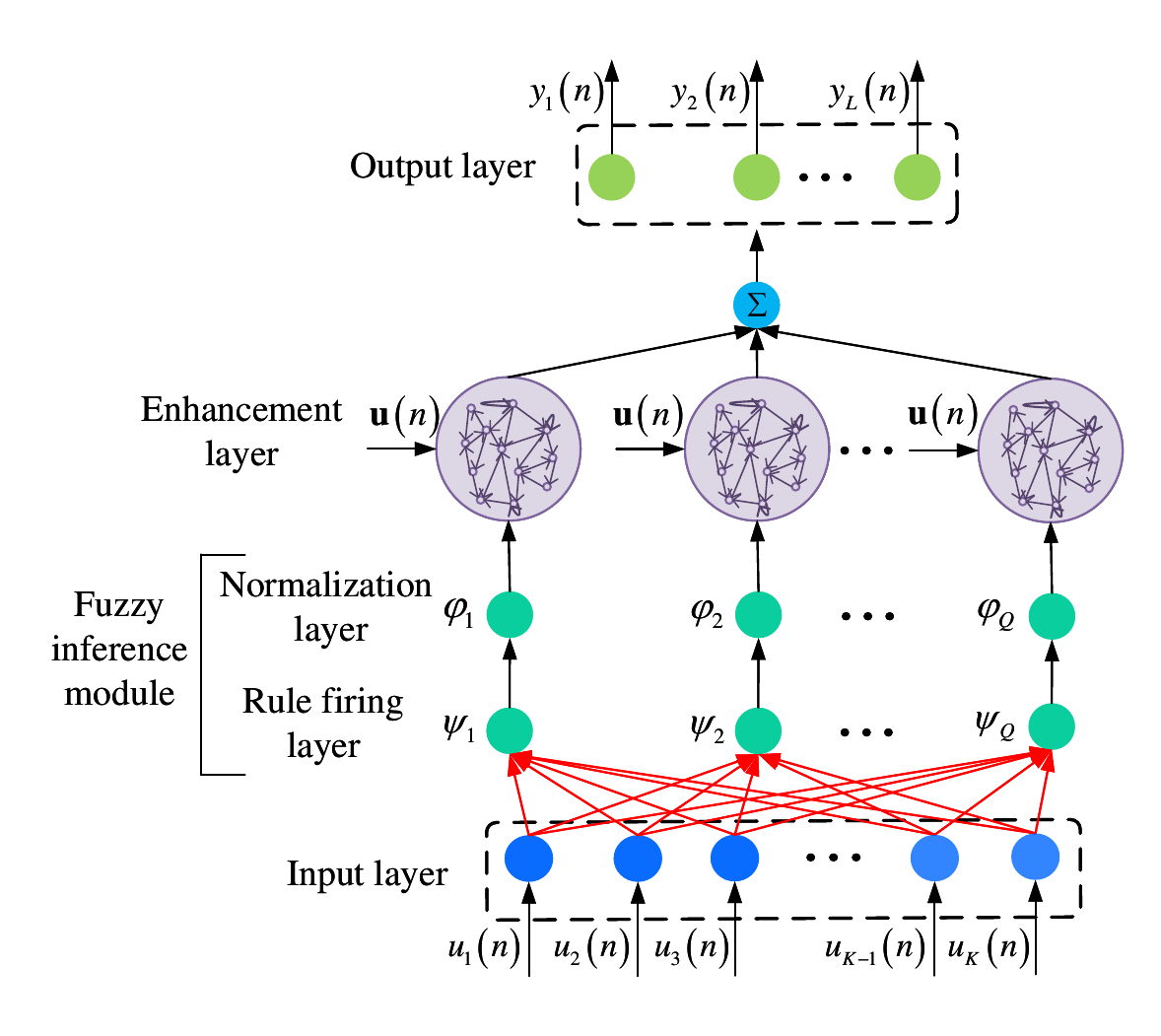}
\caption{Architecture of the fuzzy RSCN.}
\label{fig1}
\end{figure}
This section details the proposed fuzzy RSC frameworks, which integrate TSK fuzzy inference systems with RSCNs. This hybrid structure is constructed on the basic RSCN with multiple sub-reservoirs and establishes the fuzzy inference module between the input data and sub-reservoirs. Each sub-reservoir is contributed with a TSK fuzzy rule. The network structure is determined by the RSC algorithm, which ensures the built model can infinitely approximate any nonlinear maps. The architecture of the fuzzy RSCN is shown in Fig.~\ref{fig1}, and the training algorithm is summarized in Algorithm 1. Additionally, the echo state property, parameters learning and convergence analysis of fuzzy RSCNs are given in this section.

Given the dataset $\left\{ \mathbf{U},\mathbf{T} \right\}$, $\mathbf{U}=\left[ \mathbf{u}\left( 1 \right),\ldots ,\mathbf{u}\left( {{n}_{\max }} \right) \right]\in {{\mathbb{R}}^{K\times {{n}_{\max }}}}$, $\mathbf{T}=\left[ \mathbf{t}\left( 1 \right),\ldots ,\mathbf{t}\left( {{n}_{\max }} \right) \right]\in {{\mathbb{R}}^{L\times {{n}_{\max }}}}$, $\mathbf{u}\left( n \right)=\left[ {{u}_{1}}\left( n \right),\ldots ,{{u}_{K}}\left( n \right) \right]^{\top }$, $\mathbf{t}\left( n \right)=\left[ {{t}_{1}}\left( n \right),\ldots ,{{t}_{L}}\left( n \right) \right]^{\top }$. The fuzzy module with $Q$ fuzzy rules is formulated as
\begin{footnotesize}
\begin{equation}
\label{eq15}
\begin{array}{l}
{\rm{Rul}}{{\rm{e}}^i}:{\rm{If}}{\kern 1pt} {\kern 1pt} {u_1}\left( n \right){\kern 1pt} {\kern 1pt} {\rm{is}}{\kern 1pt} {\kern 1pt} {\kern 1pt} {A_i},{\kern 1pt} {\kern 1pt} {u_2}\left( n \right){\kern 1pt} {\kern 1pt} {\rm{is}}{\kern 1pt} {\kern 1pt} {A_i},{\kern 1pt} {\kern 1pt}  \ldots ,{\kern 1pt} {\kern 1pt} {\rm{and}}{\kern 1pt} {u_K}\left( n \right){\kern 1pt} {\kern 1pt} {\kern 1pt} {\rm{is}}{\kern 1pt} {\kern 1pt} {A_i},\\
{\kern 1pt} {\rm{Then}}:\left\{ {\begin{array}{*{20}{c}}
{{{\bf{x}}^i}\left( n \right) = g\left( {{\bf{W}}_{{\rm{in}}}^i{\bf{u}}(n) + {\bf{W}}_{\rm{r}}^i{{\bf{x}}^i}(n - 1) + {{\bf{b}}^i}} \right)}\\
{{{\bf{y}}^i}(n) = {\bf{W}}_{{\rm{out}}}^i\left( {{{\bf{x}}^i}\left( n \right),{\bf{u}}(n)} \right)}
\end{array}} \right.,\\
{\kern 1pt} {\kern 1pt} {\kern 1pt} {\kern 1pt} {\kern 1pt} {\kern 1pt} {\kern 1pt} {\kern 1pt} {\kern 1pt} {\kern 1pt} {\kern 1pt} {\kern 1pt} {\kern 1pt} {\kern 1pt} {\kern 1pt} {\kern 1pt} {\kern 1pt} {\kern 1pt} {\kern 1pt} {\kern 1pt} {\kern 1pt} {\kern 1pt} {\kern 1pt} {\kern 1pt} {\kern 1pt} {\kern 1pt} {\kern 1pt} {\kern 1pt} {\kern 1pt} {\kern 1pt} {\kern 1pt} {\kern 1pt} {\kern 1pt} {\kern 1pt} {\kern 1pt} {\kern 1pt} {\kern 1pt} {\kern 1pt} {\kern 1pt} {\kern 1pt} {\kern 1pt} {\kern 1pt} {\kern 1pt} {\kern 1pt} {\kern 1pt} {\kern 1pt} {\kern 1pt} {\kern 1pt} {\kern 1pt} {\kern 1pt} {\kern 1pt} {\kern 1pt} {\kern 1pt} {\kern 1pt} {\kern 1pt} {\kern 1pt} {\kern 1pt} {\kern 1pt} {\kern 1pt} {\kern 1pt} {\kern 1pt} {\kern 1pt} {\kern 1pt} {\kern 1pt} {\kern 1pt} {\kern 1pt} {\kern 1pt} {\kern 1pt} {\kern 1pt} {\kern 1pt} {\kern 1pt} {\kern 1pt} {\kern 1pt} {\kern 1pt} i = 1,2, \ldots ,Q,
\end{array}
\end{equation}
\end{footnotesize}
where $\mathbf{W}_{\text{in}}^{i},\mathbf{W}_{\text{r}}^{i},{{\mathbf{b}}^{i}}$, and $\mathbf{W}_{\text{out}}^{i}$ are the input weight matrix, reservoir weight matrix, input bias, and output weight matrix of the $i\text{-th}$ sub-reservoir, respectively. ${{\mathbf{x}}^{i}}\left( n \right)$ and ${{\mathbf{y}}^{i}}(n)$ denote the reservoir state and output of the $i\text{-th}$ sub-reservoir.

\subsection{Construction of fuzzy RSCNs}
As shown in Fig.~\ref{fig1}, fuzzy RSCNs consist of five layers, namely the input layer, rule firing layer, normalization layer, enhancement layer, and output layer. The functions of each layer are described in this subsection.

\subsubsection{Input layer}
The input layer is the first layer in the network, which is responsible for receiving and processing the raw input data so that the subsequent layers can better learn the input information. The number of nodes in this layer is the dimension of the input data. Assume the $n\text{-th}$ sample is employed to train the network, then, the input is $\mathbf{u}\left( n \right)=\left[ {{u}_{1}}\left( n \right),\ldots ,{{u}_{K}}\left( n \right) \right]^{\top }\in {{\mathbb{R}}^{K}}$.

\subsubsection{Rule firing layer}
The second layer is the rule firing layer and each node in this layer corresponds to a fuzzy rule. The fire strength reflects the degree to which associated rule conditions are satisfied by the input information. The fuzzy membership function can be used for calculating the fire strength ${{\psi }_{i}}$ of the $i\text{-th}$ fuzzy rule, represented by a Gaussian function as
\begin{equation}
\label{eq16}
{{\psi }_{i}}=\prod\limits_{k=1}^{K}{\exp \left( -{{\left( \frac{{{u}_{k}}-c_{k}^{i}}{\sigma _{k}^{i}} \right)}^{2}} \right)},i=1,2,\ldots ,Q,
\end{equation}
where $c_{k}^{i}$ and $\sigma _{k}^{i}$ are the center and width of the $k\text{-th}$ input for the $i\text{-th}$ fuzzy rule, respectively.

\subsubsection{Normalization layer}
The third layer is the normalization layer, whose function is to adjust the distribution of $\left\{ {{\psi }_{1}},{{\psi }_{2}},\ldots ,{{\psi }_{Q}} \right\}$, and improve the stability and training speed of the model. The fire strength of each fuzzy rule given in the last layer is normalized by
\begin{equation}
\label{eq17}
{{\varphi }_{i}}=\frac{{{\psi }_{i}}}{\sum\limits_{i=1}^{Q}{{{\psi }_{i}}}}.
\end{equation}

\subsubsection{Enhancement layer}
The fourth layer is the enhancement layer, where multiple sub-reservoirs are used to process the dynamic nonlinear data. Each sub-reservoir is constructed by using the RSC algorithm, which guarantees the universal approximation capability of the network. The state and output of the $i\text{-th}$ sub-reservoir with $N$ reservoir nodes can be calculated by
\begin{equation}
\label{eq18}
\mathbf{x}_{N}^{i}\left( n \right)=g\left( \mathbf{W}_{\text{in,}N}^{i}\mathbf{u}(n)+\mathbf{W}_{\text{r,}N}^{i}\mathbf{x}_{N}^{i}(n-1)+\mathbf{b}_{N}^{i} \right),
\end{equation}
\begin{equation}
\label{eq19}
{{\mathbf{y}}^{i}}(n)=\mathbf{W}_{\text{out,}N}^{i}\left( \mathbf{x}_{N}^{i}\left( n \right),\mathbf{u}(n) \right),
\end{equation}
where $\mathbf{W}_{\text{in,}N}^{i},\mathbf{W}_{\text{r,}N}^{i},\mathbf{b}_{N}^{i}$, $\mathbf{W}_{\text{out,}N}^{i}$, and $\mathbf{x}_{N}^{i}\left( n \right)$ are the input weight matrix, reservoir weight matrix, input bias, output weight matrix, and reservoir state of the $i\text{-th}$ sub-reservoir with $N$ reservoir nodes. The residual error is denoted as $e_{N}^{i}={{\mathbf{Y}}^{i}}-\mathbf{T}$, where ${{\mathbf{Y}}^{i}}=\left[ {{\mathbf{y}}^{i}}\left( 1 \right),{{\mathbf{y}}^{i}}\left( 2 \right),\ldots ,{{\mathbf{y}}^{i}}\left( {{n}_{\max }} \right) \right]$ and $\mathbf{T}=\left[ \mathbf{t}\left( 1 \right),\mathbf{t}\left( 2 \right),...\mathbf{t}\left( {{n}_{max}} \right) \right]$ are the model output of the $i\text{-th}$ sub-reservoir and the desired output, respectively. Once $e_{N}^{i}$ cannot satisfy the preset error threshold, it is necessary to add nodes under the supervisory mechanism. The configuration process can be summarized as follows.\\
\textbf{Step 1:} Generate $\left[ {w_{{\mathop{\rm i}\nolimits} {\mathop{\rm n}\nolimits} }^{N + 1,1},w_{{\mathop{\rm i}\nolimits} {\mathop{\rm n}\nolimits} }^{N + 1,2}, \cdots ,w_{{\mathop{\rm i}\nolimits} {\mathop{\rm n}\nolimits} }^{N + 1,K}} \right]$, $\left[ w_{\operatorname{r}}^{N\text{+1,1}},w_{\operatorname{r}}^{N\text{+1,2}},\ldots ,w_{\operatorname{r}}^{N\text{+}1,N+1} \right]$, and ${{b}_{N+1}}$ stochastically in ${{G}_{\max }}$ times from the adjustable uniform distribution $\left[ -\lambda ,\lambda  \right]$, $\lambda \in \left\{ {{\lambda }_{\min }},{{\lambda }_{\min }}+\Delta \lambda ,...,{{\lambda }_{\max }} \right\}$. Construct the reservoir weight matrix with a special structure
\begin{small}
\begin{equation}
\label{eq21}
\mathbf{W}_{\operatorname{r},N\text{+}1}^{i}\text{=}\left[ \begin{matrix}
   w_{\operatorname{r}}^{1,1} & 0 & \cdots  & 0 & 0  \\
   w_{\operatorname{r}}^{2,1} & w_{\operatorname{r}}^{2,2} & \cdots  & 0 & 0  \\
   \vdots  & \vdots  & \vdots  & \vdots  & \vdots   \\
   w_{\operatorname{r}}^{N,1} & w_{\operatorname{r}}^{N,2} & \cdots  & w_{\operatorname{r}}^{N,N} & 0  \\
   w_{\operatorname{r}}^{N\text{+1,1}} & w_{\operatorname{r}}^{N\text{+1,2}} & \cdots  & w_{\operatorname{r}}^{N\text{+}1,N} & w_{\operatorname{r}}^{N\text{+}1,N+1}  \\
\end{matrix} \right].
\end{equation}
\end{small}

\begin{remark}
To ensure the echo state property of the network, ${\bf{W}}_{\operatorname{r},N{\rm{ + }}1}^i$ should be reset as
\begin{equation}
\label{eq211}
{\bf{W}}_{\operatorname{r},N{\rm{ + }}1}^i{\rm{ = }}\frac{\alpha }{{\rho \left( {{\bf{W}}_{\operatorname{r},N{\rm{ + }}1}^i} \right)}}{\bf{W}}_{\operatorname{r},N{\rm{ + }}1}^i,
\end{equation}
where $0 < \alpha  < 1$ is the scaling factor, and ${\rho \left( {{\bf{W}}_{r,N{\rm{ + }}1}^i} \right)}$ is the maximum eigenvalue value of ${\bf{W}}_{r,N{\rm{ + }}1}^i$.
\end{remark}
\textbf{Step 2:} Seek the random basis function ${g_{N{\rm{ + }}1}}$ satisfying the following inequality constraint with an increasing factor ${{r}_{m}}$, $m=1,2,\ldots ,t$, $0<{{r}_{1}}<{{r}_{2}}<\ldots <{{r}_{t}}<1$,
\begin{equation}
\label{eq22}
{{\left\langle e_{N,q}^{i},{{g}_{N\text{+}1}} \right\rangle }^{2}}\ge b_{g}^{2}(1-{{r}_{m}}-{{\mu }_{N+1}})\left\| e_{N,q}^{i} \right\|^{2},q=1,2,...L,
\end{equation}
where $\left\{ {{\mu }_{N+1}} \right\}$ is a non-negative real sequence satisfying $\underset{N\to \infty }{\mathop{\lim }}\,{{\mu }_{N+1}}=0$, and ${{\mu }_{N+1}}\le (1-r)$, $L$ is the number of output layer nodes, and $0<{{\left\| g \right\|}}<{{b}_{g}}$.\\
\textbf{Step 3:} Define a set of variables $\left[ \xi _{N+1,1}^{i},\xi _{N+1,2}^{i},...,\xi _{N+1,L}^{i} \right]$
\begin{equation}
\label{eq23}
\xi _{N+1,q}^{i}=\frac{{{\left( e_{N,q}^{i\top }{{g}_{N+1}} \right)}^{2}}}{g_{N+1}^{\top }{{g}_{N+1}}}-\left( 1-{{\mu }_{N+1}}-r \right)e_{N,q}^{i\top }e_{N,q}^{i}.
\end{equation}
Select the node with the largest $\xi _{N+1}^{i}=\sum\limits_{q=1}^{L}{\xi _{N+1,q}^{i}}$ as the optimal adding node.\\
\textbf{Step 4:} Evaluate the output weights based on the global least square method
\begin{equation}
\label{eq24}
\begin{array}{l}
{\bf{W}}_{{\rm{out}},N{\rm{ + }}1}^i{\rm{ = }}\mathop {\arg \min }\limits_{{\bf{W}}_{{\rm{out}}}^{}} \left\| {{\bf{T}} - {\bf{W}}_{{\rm{out}}}^i{\bf{X}}_{N{\rm{ + }}1}^i} \right\|^2,
\end{array}
\end{equation}
where $\mathbf{X}_{N\text{+}1}^{i}\text{=}\left[ \left( \mathbf{x}_{N\text{+}1}^{i}\left( 1 \right),\mathbf{u}\left( 1 \right) \right),\ldots ,\left( \mathbf{x}_{N\text{+}1}^{i}\left( {{n}_{\max }} \right),\mathbf{u}\left( {{n}_{\max }} \right) \right) \right]$. \\
\textbf{Step 5:} Calculate the residual error $e_{N\text{+}1}^{i}$ and update $e_{0}^{i}:=e_{N\text{+}1}^{i}$, $N=N+1$. If ${{\left\| e_{0}^{i} \right\|}_{F}}\le \varepsilon $ or $N\ge {{N}_{\max }}$, we complete the configuration, where ${{\left\| \bullet  \right\|}_{F}}$ represents the F-norm and ${{N}_{\max }}$ is the maximum size of the reservoir. Otherwise, repeat steps 1-4 until satisfying the termination conditions. 

At last, we have $\underset{N\to \infty }{\mathop{\lim }}\,{{\left\| e_{N}^{i} \right\|}_{F}}=0$.

\subsubsection{Output layer}
The last layer is the output layer, where the inputs contain the normalized fire strength ${{\varphi }_{i}}$ and the output of the enhancement layer. The final output is the weighted sum of the output of the previous layer, which can be obtained by
\begin{equation}
\label{eq25}
\mathbf{y}(n)=\sum\limits_{i=1}^{Q}{{{\varphi }_{i}}{{\mathbf{y}}^{i}}(n)}.
\end{equation}

\begin{remark}
For fuzzy RSCNs, the parameters that need to be set and learned include the fuzzy inference system and neural network parameters. Traditional neuro-fuzzy models use gradient-based and intelligent optimization algorithms to determine learning parameters. However, in complex and high-dimensional feature spaces, finding the optimal set of parameters can be computationally expensive and may require a large number of iterations. To solve these problems, we set the parameters of our proposed fuzzy RSCNs as follows:
\begin{itemize}
  \item [1)] 
  The fuzzy c-mean (FCM) clustering algorithm is applied to generate the optimum centers of each fuzzy rule.       
  \item [2)]
  The RSC algorithm is used to generate the random network parameters.
  \item [3)]
  The number of fuzzy rules $Q$ and the optimum sizes of the sub-reservoirs $N$ are determined using the grid search method.
\end{itemize}
\end{remark}

A complete algorithm description of the proposed fuzzy RSCN is given in Algorithm 1. 

\begin{algorithm}[ht]\footnotesize
\caption{F-RSC}
\KwIn{Training dataset $\left( \mathbf{U},\mathbf{T} \right)$, inputs $\mathbf{U}\text{=}\left[ \mathbf{u}(1),\ldots ,\mathbf{u}({{n}_{\max }}) \right]$, desired outputs $\mathbf{T}=\left[ \mathbf{t}\left( 1 \right),...\mathbf{t}\left( {{n}_{max}} \right) \right]$, the number of fuzzy rules $Q$, the maximum size of the sub-reservoir ${{N}_{\max }}$, training error threshold $\varepsilon $, random parameter scalars $\mathbf{\gamma }\text{=}\left\{ {{\lambda }_{\min }},{{\lambda }_{\min }}+\Delta \lambda ,...,{{\lambda }_{\max }} \right\}$, and the maximum number of stochastic configurations ${{G}_{\max }}$.} 
\KwOut{Fuzzy RSCN}
\tcp{\textbf{Phase 1: Construct fuzzy rules and calculate the fire strength}}
The input data $\mathbf{U}$ is clustered using fuzzy c-mean algorithm to obtain $Q$ clustering centers;\\
Calculate the fire strength ${{\psi }_{i}}$ for each fuzzy rule based on Eq. (\ref{eq16});\\
Normalize the fire strength $\left\{ {{\psi }_{1}},{{\psi }_{2}},\ldots ,{{\psi }_{Q}} \right\}$ for each fuzzy rule and obtain the normalized values based on Eq. (\ref{eq17});\\
\tcp{\textbf{Phase 2: Configure the parameters of the \emph{i}-th sub-reservoir}}
Randomly assign $\mathbf{W}_{\text{in,}1}^{i}$, $\mathbf{W}_{\text{r,}1}^{i}$, and $\mathbf{b}_{1}^{i}$ according to the sparsity of the reservoir from $\left[ -\lambda ,\lambda  \right]$. Calculate the model output ${{\mathbf{Y}}^{i}}$ and current error $e_{N}^{i}$, where $N=1$. Set the initial residual error $e_{0}^{i}:=e_{N}^{i}$, $0<r<1$, ${\bf{\Omega }},{\bf{D}}: = \left[ {{\kern 1pt} {\kern 1pt} {\kern 1pt} {\kern 1pt} {\kern 1pt} {\kern 1pt}} \right]$;\\
        \While{$N<{{N}_{\max }}$ AND ${{\left\| e_{0}^{i} \right\|}_{F}}>\varepsilon $}{
            \For{$\lambda \in \mathbf{\gamma }$}{
                \For{$l=1,2,\ldots ,{{G}_{\max }}$}{
                Randomly assign $\mathbf{W}_{\text{in,}N+1}^{i}$, $\mathbf{W}_{\text{r,}N+1}^{i}$, and $\mathbf{b}_{N+1}^{i}$ from $\left[ -\lambda ,\lambda  \right]$;\\
                Construct the sub-reservoir state $\mathbf{X}_{N\text{+}1}^{i}$;\\
                Calculate $\xi _{N+1,q}^{i}$ based on Eq. (\ref{eq23});\\
                \eIf{$\min \left\{ \xi _{N+1,1}^{i},\xi _{N+1,2}^{i},...,\xi _{N+1,L}^{i} \right\}\ge 0$}{
                    Save $\mathbf{W}_{\text{in,}N+1}^{i}$, $\mathbf{W}_{\text{r,}N+1}^{i}$, and $\mathbf{b}_{N+1}^{i}$ in $\mathbf{D}$, and $\xi _{N+1}^{i}=\sum\limits_{q=1}^{L}{\xi _{N+1,q}^{i}}$ in $\mathbf{\Omega }$;\\
                    }{
                Go back to Step 7;\\
                }
                }
                \eIf{$\mathbf{D}$ is not empty}{
                    Find $\mathbf{W}_{\text{in,}N+1}^{i*}$, $\mathbf{W}_{\text{r,}N+1}^{i*}$, and $\mathbf{b}_{N+1}^{i*}$ that maximize $\xi _{N+1}^{i}$ in $\mathbf{\Omega }$, and get $\mathbf{X}_{N\text{+}1}^{i*}$;\\
                    \textbf{Break} (go to Step 24;)
                    }{
                    Randomly take $\tau \in \left( 0,1-r \right)$ and update $r=r+\tau $, return to Step 7;\\
                    }   
            }
            Calculate output weights $\mathbf{W}_{\text{out},N\text{+}1}^{i*}\text{=}\left[ \mathbf{w}_{\text{out},1}^{*},...,\mathbf{w}_{\text{out},N+1+K}^{*} \right]$ based on Eq. (\ref{eq24});\\
            Calculate $e_{N\text{+}1}^{i*}\text{=}e_{N}^{i*}-\mathbf{W}_{\text{out}}^{i*}\mathbf{X}_{N\text{+}1}^{i*}$;\\
            Update $e_{0}^{i}:=e_{N\text{+}1}^{i*}$, $N=N+1$;\\
        }
        \tcp{\textbf{Phase 3: Calculate the final output}}
        Calculate the weighted sum of the output of each sub-reservoir based on Eq. (\ref{eq25});\\
        \textbf{Return} $\left\{ \mathbf{W}_{\text{out},N}^{1*},\ldots ,\mathbf{W}_{\text{out},N}^{Q*} \right\}$, $\left\{ \mathbf{W}_{\text{in,}N}^{1*},\mathbf{W}_{\text{in,}N}^{Q*} \right\}$, $\left\{ \mathbf{W}_{\text{r,}N}^{1*},\ldots ,\mathbf{W}_{\text{r,}N}^{Q*} \right\}$, and $\left\{ \mathbf{b}_{N}^{1*},\ldots ,\mathbf{b}_{N}^{Q*} \right\}$.
\end{algorithm}

\subsection{Echo state property of F-RSCNs}
The echo state property enhances the model's ability to handle intricate scenarios such as long-term dependencies, high-dimensional inputs, real-time data, and dynamic environments. This unique property allows the network to retain and echo previous input information in future time steps, thereby facilitating the analysis and processing of temporal data.\\

\textbf{Theorem 2.} Define two reservoir states ${{\bf{x}}_{N + 1}^{i,1}\left( n \right)}$ and ${{\bf{x}}_{N + 1}^{i,2}\left( n \right)}$ for the $i\text{-th}$ sub-reservoir, which are driven by the same input sequence but start from different initial conditions. The proposed fuzzy RSCN holds the echo state property if the scaling factor is chosen as 
\begin{equation}
\label{eq212}
0 < \alpha  < \frac{{\rho \left( {{\bf{W}}_{r,N{\rm{ + }}1}^i} \right)}}{{{\sigma _{\max }}\left( {{\bf{W}}_{r,N{\rm{ + }}1}^i} \right)}},
\end{equation}
where ${{\sigma _{\max }}\left( {{\bf{W}}_{r,N{\rm{ + }}1}^i} \right)}$ is the the maximal singular value of ${\bf{W}}_{r,N{\rm{ + }}1}^i$.\\
\textbf{Proof.} The error between ${{\bf{x}}_{N + 1}^{i,1}\left( n \right)}$ and ${{\bf{x}}_{N + 1}^{i,2}\left( n \right)}$ can be expressed as
\begin{footnotesize}
 \begin{equation}
\label{eq213}
\begin{array}{*{20}{l}}
{{{\left\| {{\bf{x}}_{N + 1}^{i,1}\left( n \right) - {\bf{x}}_{N + 1}^{i,2}\left( n \right)} \right\|}^2}}\\
{ = \left\| {g\left( {{\bf{W}}_{{\rm{in,}}N + 1}^i{\bf{u}}(n) + {\bf{W}}_{{\rm{r,}}N + 1}^i{\bf{x}}_{N + 1}^{i,1}\left( {n - 1} \right) + {\bf{b}}_{N + 1}^i} \right)} \right.}\\
{  - {{\left. {g\left( {{\bf{W}}_{{\rm{in,}}N + 1}^i{\bf{u}}(n) + {\bf{W}}_{{\rm{r,}}N + 1}^i{\bf{x}}_{N + 1}^{i,2}\left( {n - 1} \right) + {\bf{b}}_{N + 1}^i} \right)} \right\|}^2}}
\end{array},
\end{equation}   
\end{footnotesize}
where $g$ is selected from the sigmoid or tanh function. According to the Lipschitz condition, yields
\begin{small}
\begin{equation}
\label{eq214}
\begin{array}{*{20}{l}}
{{{\left\| {{\bf{x}}_{N + 1}^{i,1}\left( n \right) - {\bf{x}}_{N + 1}^{i,2}\left( n \right)} \right\|}^2}}\\
{ \le \max \left( {\left| {g'} \right|} \right){{\left\| {{\bf{W}}_{{\rm{r,}}N + 1}^i{\bf{x}}_{N + 1}^{i,1}\left( {n - 1} \right) - {\bf{W}}_{{\rm{r,}}N + 1}^i{\bf{x}}_{N + 1}^{i,2}\left( {n - 1} \right)} \right\|}^2}}\\
{ \le \left\| {{\bf{W}}_{{\rm{r,}}N + 1}^i} \right\|{{\left\| {{\bf{x}}_{N + 1}^{i,1}\left( {n - 1} \right) - {\bf{x}}_{N + 1}^{i,2}\left( {n - 1} \right)} \right\|}^2}}\\
{ \le \left\| {{\bf{W}}_{{\rm{r,}}N + 1}^i} \right\|_{}^2{{\left\| {{\bf{x}}_{N + 1}^{i,1}\left( {n - 2} \right) - {\bf{x}}_{N + 1}^{i,2}\left( {n - 2} \right)} \right\|}^2}}\\
{ \le  \ldots }{ \le \left\| {{\bf{W}}_{{\rm{r,}}N + 1}^i} \right\|_{}^n{{\left\| {{\bf{x}}_{N + 1}^{i,1}\left( 0 \right) - {\bf{x}}_{N + 1}^{i,2}\left( 0 \right)} \right\|}^2}}
\end{array}
\end{equation}    
\end{small}
where ${g'}$ is the first order derivative of $g$. Observe that
\begin{equation}
\label{eq215}
\begin{array}{l}
{\sigma _{\max }}\left( {\frac{\alpha }{{\rho \left( {{\bf{W}}_{r,N{\rm{ + }}1}^i} \right)}}{\bf{W}}_{r,N{\rm{ + }}1}^i} \right)\\
 = \frac{\alpha }{{\rho \left( {{\bf{W}}_{r,N{\rm{ + }}1}^i} \right)}}{\sigma _{\max }}\left( {{\bf{W}}_{r,N{\rm{ + }}1}^i} \right)\\
 < \frac{{\rho \left( {{\bf{W}}_{r,N{\rm{ + }}1}^i} \right)}}{{{\sigma _{\max }}\left( {{\bf{W}}_{r,N{\rm{ + }}1}^i} \right)}} \times \frac{1}{{\rho \left( {{\bf{W}}_{r,N{\rm{ + }}1}^i} \right)}}{\sigma _{\max }}\left( {{\bf{W}}_{r,N{\rm{ + }}1}^i} \right) = 1.
\end{array}
\end{equation}
The Euclidean norm of the matrix is equal to the largest singular value. Combining Eq. (\ref{eq211}) and Eq. (\ref{eq215}), we have $\left\| {{\bf{W}}_{r,N{\rm{ + }}1}^i} \right\| < 1$ and $\mathop {\lim }\limits_{n \to \infty } {\mkern 1mu} \left\| {{\bf{W}}_{r,N{\rm{ + }}1}^i} \right\|_{}^n = 0$. It can be easily inferred that $\mathop {\lim }\limits_{n \to \infty } {\mkern 1mu} \left\| {{\bf{x}}_{N + 1}^{i,1}\left( n \right) - {\bf{x}}_{N + 1}^{i,2}\left( n \right)} \right\| = 0$, which completes the proof.

\subsection{Universal approximation property of F-RSCNs}
Fuzzy RSCNs have the ability to approximate any complex functions, offering a significant advantage in modelling uncertain complex industrial systems. This subsection provides a detailed theoretical proof of the universal approximation property for the proposed approach.\\

\textbf{Theorem 3.} Suppose that $span\left( \Gamma  \right)$ is dense in ${{L}_{2}}$ space. For $b_{g}^{*}\in {{\mathbb{R}}^{+}}$, $\forall {g}\in \Gamma$, $0<{{\left\| g \right\|}}<b_{g}^{*}$. Given $0<r<1$ and a nonnegative real sequence $\left\{ {{\mu }_{N+1}} \right\}$ satisfying $\underset{N\to \infty }{\mathop{\lim }}\,{{\mu }_{N+1}}=0$, and ${{\mu }_{N+1}}\le (1-r)$. For $N=1,2...$, $q=1,2,...,L$, and $i=1,2,...,Q$, define
\begin{equation}
\label{eq26}
\delta _{N\text{+}1}^{i}=\sum\limits_{q=1}^{L}{\delta _{N\text{+}1,q}^{i}},\delta _{N\text{+}1,q}^{i}=(1-r-{{\mu }_{N\text{+}1}})\left\| e_{N,q}^{i} \right\|^{2}.
\end{equation}
If ${{g}_{N+1}}$ satisfies the following inequality constraints
\begin{equation}
\label{eq27}
{{\left\langle e_{N,q}^{i},{{g}_{N\text{+}1}} \right\rangle }^{2}}\ge b_{g}^{*2}\delta _{N\text{+}1,q}^{i},q=1,2,\ldots L,
\end{equation}
and the output weight is constructively evaluated by
\begin{equation}
\label{eq28}
\mathbf{w}_{\operatorname{out},N+1,q}^{i*}=\frac{\left\langle e_{N,q}^{i},g_{N\text{+}1}^{{}} \right\rangle }{\left\| g_{N\text{+}1}^{{}} \right\|^{2}}.
\end{equation}
Then, we have $\underset{N\to \infty }{\mathop{\lim }}\,\left\| e_{N+1}^{i} \right\|=0$.\\
\textbf{Proof.} Note that
\begin{small}
\begin{equation}
\label{eq29}
\begin{array}{l}
\left\| {e_{N{\rm{ + }}1}^i} \right\|^2 - \left( {r + {\mu _{N{\rm{ + }}1}}} \right)\left\| {e_N^i} \right\|^2\\
{\kern 1pt}  = \sum\limits_{q = 1}^L {\left\langle {e_{N,q}^i - {\bf{w}}_{{\mathop{\rm out}\nolimits} ,N + 1,q}^{i * }g_{N + 1}^{},} \right.} {\kern 1pt} {\kern 1pt} \left. {{\kern 1pt} {\kern 1pt} e_{N,q}^i - {\bf{w}}_{{\mathop{\rm out}\nolimits} ,N + 1,q}^{i * }g_{N + 1}^{}} \right\rangle \\
{\kern 1pt} {\kern 1pt} {\kern 1pt} {\kern 1pt} {\kern 1pt} {\kern 1pt} {\kern 1pt} {\kern 1pt} {\kern 1pt} {\kern 1pt} {\kern 1pt} {\kern 1pt} {\kern 1pt} {\kern 1pt} {\kern 1pt} {\kern 1pt} {\kern 1pt} {\kern 1pt} {\kern 1pt} {\kern 1pt} {\kern 1pt} {\kern 1pt} {\kern 1pt} {\kern 1pt} {\kern 1pt} {\kern 1pt} {\kern 1pt} {\kern 1pt} {\kern 1pt} {\kern 1pt} {\kern 1pt} {\kern 1pt} {\kern 1pt} {\kern 1pt}  - \sum\limits_{q = 1}^L {\left( {r + {\mu _{N{\rm{ + }}1}}} \right)\left\langle {e_{N,q}^i,e_{N,q}^i} \right\rangle } \\
 = \left( {1 - r - {\mu _{N{\rm{ + }}1}}} \right)\left\| {e_N^i} \right\|^2 - \sum\limits_{q = 1}^L {\left( {2\left\langle {e_{N,q}^i,{\bf{w}}_{{\mathop{\rm out}\nolimits} ,N + 1,q}^{i * }g_{N + 1}^{}} \right\rangle } \right.} \\
{\kern 1pt} {\kern 1pt} {\kern 1pt} {\kern 1pt} {\kern 1pt} {\kern 1pt} {\kern 1pt} {\kern 1pt} {\kern 1pt} {\kern 1pt} {\kern 1pt} {\kern 1pt} {\kern 1pt} {\kern 1pt} {\kern 1pt} {\kern 1pt} {\kern 1pt} {\kern 1pt} {\kern 1pt} {\kern 1pt} {\kern 1pt} {\kern 1pt} {\kern 1pt} {\kern 1pt} {\kern 1pt} {\kern 1pt} {\kern 1pt} {\kern 1pt} {\kern 1pt} {\kern 1pt} {\kern 1pt} {\kern 1pt} {\kern 1pt} {\kern 1pt} {\kern 1pt} {\kern 1pt} \left. { - \left\langle {{\bf{w}}_{{\mathop{\rm out}\nolimits} ,N + 1,q}^{i * }g_{N + 1}^{},{\bf{w}}_{{\mathop{\rm out}\nolimits} ,N + 1,q}^{i * }g_{N + 1}^{}} \right\rangle } \right)\\
 = \left( {1 - r - {\mu _{N{\rm{ + }}1}}} \right)\left\| {e_N^i} \right\|_2^2 - \frac{{\sum\limits_{q = 1}^L {{{\left\langle {e_{N,q}^i,g_{N + 1}^{}} \right\rangle }^2}} }}{{\left\| {g_{N{\rm{ + }}1}^{}} \right\|^2}}\\
 = \delta _{N{\rm{ + }}1}^i - \frac{{\sum\limits_{q = 1}^L {{{\left\langle {e_{N,q}^i,g_{N + 1}^{}} \right\rangle }^2}} }}{{\left\| {g_{N{\rm{ + }}1}^{}} \right\|^2}}\le \delta _{N{\rm{ + }}1}^i - \frac{{\sum\limits_{q = 1}^L {{{\left\langle {e_{N,q}^i,g_{N + 1}^{}} \right\rangle }^2}} }}{{b_g^{ * 2}}} \le 0.
\end{array}
\end{equation}
\end{small}
Thus, we have $\left\| e_{N\text{+}1}^{i} \right\|^{2}-\left( r+{{\mu }_{N\text{+}1}} \right)\left\| e_{N}^{i} \right\|^{2}\le 0$. Notably, the global least squares method is used to calculate the output weight in our proposed algorithm. We can obtain the optimum output weight $\mathbf{W}_{\text{out},N\text{+}1}^{i}\text{=}\left[ \mathbf{w}_{\text{out},1}^{i},\mathbf{w}_{\text{out},2}^{i},...,\mathbf{w}_{\text{out},N+1+K}^{i} \right]$ and residual error $\tilde{e}_{N\text{+}1}^{i}$ based on Eq. (24). It is easily inferred that $\left\| \tilde{e}_{N\text{+}1}^{i} \right\|^{2}\le \left\| e_{N\text{+}1}^{i} \right\|^{2}$. Therefore, the following inequalities can be established
\begin{small}
 \begin{equation}
\label{eq30}
\left\| \tilde{e}_{N\text{+}1}^{i} \right\|^{2}\le r\left\| e_{N}^{i} \right\|^{2}+\gamma _{N+1}^{i},\left( \gamma _{N+1}^{i}={{\mu }_{N\text{+}1}}\left\| e_{N}^{i} \right\|^{2}\ge 0 \right).
\end{equation}   
\end{small}
Note that $\underset{N\to \infty }{\mathop{\lim }}\,\gamma _{N+1}^{i}=0$. Obviously, $\underset{N\to \infty }{\mathop{\lim }}\,\left\| \tilde{e}_{N+1}^{i} \right\|^{2}=0$, which implies $\underset{N\to \infty }{\mathop{\lim }}\,\left\| \tilde{e}_{N+1}^{i} \right\|=0$. This completes the proof.

\subsection{Parameter learning of F-RSCNs}
Due to the dynamic changes in the process environment, we need to update learning parameters to cope with various complex uncertainties. In this subsection, the projection algorithm \cite{ref33} is used to update the output weights online. Combining Eq. (\ref{eq19}) and Eq. (\ref{eq25}), the model output can be written as
\begin{equation}
\label{eq31}
\mathbf{y}(n)=\mathbf{W}_{\text{out}}^{1}{{\varphi }_{1}}{{\mathbf{g}}^{1}}\left( n \right)+\ldots .+\mathbf{W}_{\text{out}}^{Q}{{\varphi }_{Q}}{{\mathbf{g}}^{Q}}\left( n \right),
\end{equation}
where $\mathbf{W}_{\text{out}}^{i}$ is the output weight of the \emph{i}-th sub-reservoir and ${{\mathbf{g}}^{i}}\left( n \right)=\left( \mathbf{x}_{N}^{i}\left( n \right),\mathbf{u}(n) \right)$. Define $\Theta =\left[ \mathbf{W}_{\text{out}}^{1},\ldots ,\mathbf{W}_{\text{out}}^{Q} \right]$ and $\mathbf{G}\left( n \right)=\left[ {{\varphi }_{1}}{{\mathbf{g}}^{1}}\left( n \right),\ldots ,{{\varphi }_{Q}}{{\mathbf{g}}^{Q}}\left( n \right) \right]^{\top }$, we have
\begin{equation}
\label{eq32}
\mathbf{y}(n)=\Theta \mathbf{G}\left( n \right).
\end{equation}
A nonlinear function $f\left( x \right):{{\mathbb{R}}^{K}}\to {{\mathbb{R}}^{L}}$ on a large compact set ${{\mathbf{\Omega }}^{*}}\in {{\mathbb{R}}^{L}}$ can be approximated by the proposed fuzzy RSCN in Eq. (\ref{eq32}), so that the following inequality holds
\begin{equation}
\label{eq33}
{{\sup }_{x\in {{\mathbf{\Omega }}^{*}}}}\left| f\left( x \right)-\mathbf{y}(x) \right|\le {{\eta }_{m}},
\end{equation}
where ${{\eta }_{m}}>0$. $f\left( x \right)$ can be represented by
\begin{equation}
\label{eq34}
f\left( x \right)={{\Theta }^{*}}\mathbf{G}\left( x \right)+\eta ,\forall x\in {{\mathbf{\Omega }}^{*}},
\end{equation}
where $\eta $ is the approximation error bounded by $\left| \eta  \right|\le {{\eta }_{m}}$, and ${{\Theta }^{*}}$ is the ideal output weights satisfying
\begin{equation}
\label{eq35}
{{\Theta }^{*}}=\arg \underset{\Theta }{\mathop{\min }}\,\left\{ {{\sup }_{x\in {{\mathbf{\Omega }}^{*}}}}\left| f\left( x \right)-\Theta \mathbf{G}\left( x \right) \right| \right\}.
\end{equation}
In practice, the value of ${{\Theta }^{*}}$ is unknown, and its estimation is employed and updated through the learning algorithm to minimize the approximation error asymptotically. 

According to Eq. (\ref{eq14}), we can obtain the least squares solution for the output weights
\begin{equation}
\label{eq36}
{{\Theta }^{\top }}={{\left( \Sigma {{\Sigma }^{\top }} \right)}^{-1}}\Sigma {{\mathbf{T}}^{\top }},
\end{equation}
where $\Sigma \left( {{n}_{\max }} \right)=\left[ \mathbf{G}\left( 1 \right),\ldots ,\mathbf{G}\left( {{n}_{\max }} \right) \right]$, $\mathbf{T}\left( {{n}_{max}} \right)=\left[ \mathbf{t}\left( 1 \right),...\mathbf{t}\left( {{n}_{max}} \right) \right]$. Define
\begin{small}
 \begin{equation}
\label{eq37}
\mathbf{H}\left( n \right)={{\left( \Sigma \left( n \right){{\Sigma }^{\top }}\left( n \right) \right)}^{-1}}={{\left( \sum\limits_{j=1}^{n}{\mathbf{G}\left( j \right){{\mathbf{G}}^{\top }}\left( j \right)} \right)}^{-1}},
\end{equation}   
\end{small}
and $\Theta $ in Eq. (\ref{eq36}) at $n$ time step is denoted as
\begin{small}
 \begin{equation}
\label{eq39}
{{\Theta }^{\top }}\left( n \right)=\mathbf{H}\left( n \right)\Sigma \left( n \right){{\mathbf{T}}^{\top }}\left( n \right)=\mathbf{H}\left( n \right)\left( \sum\limits_{j=1}^{n}{\mathbf{G}\left( j \right){{\mathbf{t}}^{\top }}\left( j \right)} \right).
\end{equation}   
\end{small}
Then, we have
\begin{equation}
\label{eq40}
{{\mathbf{H}}^{-1}}\left( n \right)={{\mathbf{H}}^{-1}}\left( n-1 \right)+\mathbf{G}\left( n \right){{\mathbf{G}}^{\top }}\left( n \right),
\end{equation}
and
\begin{equation}
\label{eq41}
\begin{array}{l}
{{\bf{H}}^{ - 1}}\left( {n - 1} \right){\Theta ^ \top }\left( {n - 1} \right)\\
 = {{\bf{H}}^{ - 1}}\left( {n - 1} \right){\bf{H}}\left( {n - 1} \right)\left( {\sum\limits_{j = 1}^{n - 1} {{\bf{G}}\left( j \right){{\bf{t}}^ \top }\left( j \right)} } \right)\\
 = \sum\limits_{j = 1}^{n - 1} {{\bf{G}}\left( j \right){{\bf{t}}^ \top }\left( j \right)} .
\end{array}
\end{equation}
Substituting Eq. (\ref{eq41}) into Eq. (\ref{eq39}), yields
\begin{equation}
\label{eq42}
{{\Theta }^{\top }}\left( n \right)=\mathbf{H}\left( n \right)\left( {{\mathbf{H}}^{-1}}\left( n-1 \right){{\Theta }^{\top }}\left( n-1 \right)+\mathbf{G}\left( n \right){{\mathbf{t}}^{\top }}\left( n \right) \right).
\end{equation}
Substituting Eq. (\ref{eq40}) into Eq. (\ref{eq42}), the online update of the output weights is given by
\begin{equation}
\label{eq43}
\begin{array}{l}
{\Theta ^ \top }\left( n \right) = {\bf{H}}\left( n \right)\left( {\left( {{{\bf{H}}^{ - 1}}\left( n \right) - {\bf{G}}\left( n \right){{\bf{G}}^ \top }\left( n \right)} \right){\Theta ^ \top }\left( {n - 1} \right)} \right.\\
{\kern 1pt} {\kern 1pt} {\kern 1pt} {\kern 1pt} {\kern 1pt} {\kern 1pt} {\kern 1pt} {\kern 1pt} {\kern 1pt} {\kern 1pt} {\kern 1pt} {\kern 1pt} {\kern 1pt} {\kern 1pt} {\kern 1pt} {\kern 1pt} {\kern 1pt} {\kern 1pt} {\kern 1pt} {\kern 1pt} {\kern 1pt} {\kern 1pt} {\kern 1pt} {\kern 1pt} {\kern 1pt} {\kern 1pt} {\kern 1pt} {\kern 1pt} {\kern 1pt} {\kern 1pt} {\kern 1pt} {\kern 1pt} {\kern 1pt} {\kern 1pt} {\kern 1pt} {\kern 1pt} {\kern 1pt} {\kern 1pt} {\kern 1pt} {\kern 1pt} {\kern 1pt} {\kern 1pt} {\kern 1pt}  + \left. {{\bf{G}}\left( n \right){{\bf{t}}^ \top }\left( n \right)} \right)\\
{\kern 1pt} {\kern 1pt} {\kern 1pt}{\kern 1pt} {\kern 1pt} {\kern 1pt} {\kern 1pt} {\kern 1pt} {\kern 1pt} {\kern 1pt} {\kern 1pt} {\kern 1pt} {\kern 1pt} {\kern 1pt} {\kern 1pt} {\kern 1pt} {\kern 1pt} {\kern 1pt} {\kern 1pt} {\kern 1pt} {\kern 1pt} {\kern 1pt} {\kern 1pt} {\kern 1pt} {\kern 1pt} {\kern 1pt} {\kern 1pt} {\kern 1pt} {\kern 1pt} {\kern 1pt}  = {\bf{H}}\left( n \right)\left( {{{\bf{H}}^{ - 1}}\left( n \right){\Theta ^ \top }\left( {n - 1} \right)} \right.\\
{\kern 1pt} {\kern 1pt} {\kern 1pt} {\kern 1pt} {\kern 1pt} {\kern 1pt} {\kern 1pt} {\kern 1pt} {\kern 1pt} {\kern 1pt} {\kern 1pt} {\kern 1pt} {\kern 1pt} {\kern 1pt} {\kern 1pt} {\kern 1pt} {\kern 1pt} {\kern 1pt} {\kern 1pt} {\kern 1pt} {\kern 1pt} {\kern 1pt} {\kern 1pt} {\kern 1pt} {\kern 1pt} {\kern 1pt} {\kern 1pt} {\kern 1pt} {\kern 1pt} {\kern 1pt} {\kern 1pt} {\kern 1pt} {\kern 1pt} {\kern 1pt} {\kern 1pt} {\kern 1pt} {\kern 1pt} {\kern 1pt} {\kern 1pt} {\kern 1pt} {\kern 1pt} {\kern 1pt} {\kern 1pt}  \left. { - {\bf{G}}\left( n \right){{\bf{G}}^ \top }\left( n \right){\Theta ^ \top }\left( {n - 1} \right) + {\bf{G}}\left( n \right){{\bf{t}}^ \top }\left( n \right)} \right)\\
{\kern 1pt} {\kern 1pt} {\kern 1pt}{\kern 1pt} {\kern 1pt} {\kern 1pt} {\kern 1pt} {\kern 1pt} {\kern 1pt} {\kern 1pt} {\kern 1pt} {\kern 1pt} {\kern 1pt} {\kern 1pt} {\kern 1pt} {\kern 1pt} {\kern 1pt} {\kern 1pt} {\kern 1pt} {\kern 1pt} {\kern 1pt} {\kern 1pt} {\kern 1pt} {\kern 1pt} {\kern 1pt} {\kern 1pt} {\kern 1pt} {\kern 1pt} {\kern 1pt} {\kern 1pt}  = {\Theta ^ \top }\left( {n - 1} \right)\\
{\kern 1pt} {\kern 1pt} {\kern 1pt} {\kern 1pt} {\kern 1pt} {\kern 1pt} {\kern 1pt} {\kern 1pt} {\kern 1pt} {\kern 1pt} {\kern 1pt} {\kern 1pt} {\kern 1pt} {\kern 1pt} {\kern 1pt} {\kern 1pt} {\kern 1pt} {\kern 1pt} {\kern 1pt} {\kern 1pt} {\kern 1pt} {\kern 1pt} {\kern 1pt} {\kern 1pt} {\kern 1pt} {\kern 1pt} {\kern 1pt} {\kern 1pt} {\kern 1pt} {\kern 1pt} {\kern 1pt} {\kern 1pt} {\kern 1pt} {\kern 1pt} {\kern 1pt} {\kern 1pt} {\kern 1pt} {\kern 1pt} {\kern 1pt} {\kern 1pt} {\kern 1pt} {\kern 1pt} {\kern 1pt}  + {\bf{H}}\left( n \right){\bf{G}}\left( n \right)\left( {{{\bf{t}}^ \top }\left( n \right) - {{\bf{G}}^ \top }\left( n \right){\Theta ^ \top }\left( {n - 1} \right)} \right)\\
{\kern 1pt} {\kern 1pt} {\kern 1pt}{\kern 1pt} {\kern 1pt} {\kern 1pt} {\kern 1pt} {\kern 1pt} {\kern 1pt} {\kern 1pt} {\kern 1pt} {\kern 1pt} {\kern 1pt} {\kern 1pt} {\kern 1pt} {\kern 1pt} {\kern 1pt} {\kern 1pt} {\kern 1pt} {\kern 1pt} {\kern 1pt} {\kern 1pt} {\kern 1pt} {\kern 1pt} {\kern 1pt} {\kern 1pt} {\kern 1pt} {\kern 1pt} {\kern 1pt} {\kern 1pt}  = {\Theta ^ \top }\left( {n - 1} \right) + {\bf{H}}\left( n \right){\bf{G}}\left( n \right){\bf{e}}_s^ \top \left( n \right)\\
{\kern 1pt} {\kern 1pt} {\kern 1pt}{\kern 1pt} {\kern 1pt} {\kern 1pt} {\kern 1pt} {\kern 1pt} {\kern 1pt} {\kern 1pt} {\kern 1pt} {\kern 1pt} {\kern 1pt} {\kern 1pt} {\kern 1pt} {\kern 1pt} {\kern 1pt} {\kern 1pt} {\kern 1pt} {\kern 1pt} {\kern 1pt} {\kern 1pt} {\kern 1pt} {\kern 1pt} {\kern 1pt} {\kern 1pt} {\kern 1pt} {\kern 1pt} {\kern 1pt} {\kern 1pt}   = {\Theta ^ \top }\left( {n - 1} \right) + \frac{{{\bf{G}}\left( n \right){\bf{e}}_s^ \top \left( n \right)}}{{\sum\limits_{j = 1}^n {{\bf{G}}\left( j \right){{\bf{G}}^ \top }\left( j \right)} }},
\end{array}
\end{equation}
where $\mathbf{e}_{s}^{{}}\left( n \right)=\mathbf{t}\left( n \right)-\Theta \left( n-1 \right)\mathbf{G}\left( n \right)$. 
\begin{remark}
To prevent division by zero, a small constant \emph{c} is introduced to the denominator, and a coefficient factor \emph{a} is applied to the numerator, obtaining an improved projection algorithm 
\begin{equation}
\label{eq44}
{{\Theta }^{\top }}\left( n \right)={{\Theta }^{\top }}\left( n-1 \right)+\frac{a\mathbf{G}\left( n \right)\mathbf{e}_{s}^{\top }\left( n \right)}{c+\sum\limits_{j=1}^{n}{\mathbf{G}\left( j \right){{\mathbf{G}}^{\top }}\left( j \right)}},
\end{equation}
where $0<a\le 1$ and $c>0$.
\end{remark}

\subsection{Convergence analysis}
Convergence is crucial in industrial data analytics to ensure the algorithm can reach a stable solution over iterations. By referring to the work reported in \cite{ref29}, we give a convergence analysis on the model's parameters as follows.

At time step \emph{n}, the approximation error can be expressed as 
\begin{equation}
\label{eq45}
\begin{array}{l}
{\bf{e}}_m^{}\left( n \right) = {\bf{t}}\left( n \right) - {\bf{y}}(n) = {\Theta ^ * }{\bf{G}}\left( n \right) - \Theta \left( n \right){\bf{G}}\left( n \right)\\
{\kern 1pt} {\kern 1pt} {\kern 1pt} {\kern 1pt} {\kern 1pt} {\kern 1pt} {\kern 1pt} {\kern 1pt} {\kern 1pt} {\kern 1pt} {\kern 1pt} {\kern 1pt} {\kern 1pt} {\kern 1pt} {\kern 1pt} {\kern 1pt}{\kern 1pt} {\kern 1pt} {\kern 1pt} {\kern 1pt} {\kern 1pt} {\kern 1pt} {\kern 1pt} {\kern 1pt} {\kern 1pt} {\kern 1pt} {\kern 1pt} {\kern 1pt}  = \tilde \Theta \left( n \right){\bf{G}}\left( n \right),
\end{array}
\end{equation}
and $\mathbf{e}_{s}^{{}}\left( n \right)$ in Eq. (\ref{eq43}) can be reformulated as
\begin{equation}
\label{eq46}
\mathbf{e}_{s}^{{}}\left( n \right)={{\Theta }^{*}}\mathbf{G}\left( n \right)-\Theta \left( n-1 \right)\mathbf{G}\left( n \right)=\tilde{\Theta }\left( n-1 \right)\mathbf{G}\left( n \right),
\end{equation}
where $\tilde{\Theta }\left( n \right)={{\Theta }^{*}}-\Theta \left( n \right)$. Combining Eq. (\ref{eq43}) and Eq. (\ref{eq46}), we have
\begin{equation}
\label{eq47}
\begin{array}{l}
{\Theta ^ \top }\left( n \right) = {\Theta ^{ *  \top }} - {{\tilde \Theta }^ \top }\left( n \right)\\
{\kern 1pt} {\kern 1pt} {\kern 1pt}{\kern 1pt} {\kern 1pt} {\kern 1pt} {\kern 1pt} {\kern 1pt} {\kern 1pt} {\kern 1pt} {\kern 1pt}  {\kern 1pt} {\kern 1pt} {\kern 1pt} {\kern 1pt} {\kern 1pt} {\kern 1pt} {\kern 1pt} {\kern 1pt} {\kern 1pt} {\kern 1pt} {\kern 1pt} {\kern 1pt} {\kern 1pt} {\kern 1pt} {\kern 1pt} {\kern 1pt} {\kern 1pt} {\kern 1pt} {\kern 1pt}  = {\Theta ^ \top }\left( {n - 1} \right) + {\bf{H}}\left( n \right){\bf{G}}\left( n \right){\bf{e}}_s^ \top \left( n \right)\\
{\kern 1pt} {\kern 1pt} {\kern 1pt}{\kern 1pt} {\kern 1pt} {\kern 1pt} {\kern 1pt} {\kern 1pt} {\kern 1pt} {\kern 1pt} {\kern 1pt}  {\kern 1pt} {\kern 1pt} {\kern 1pt} {\kern 1pt} {\kern 1pt} {\kern 1pt} {\kern 1pt} {\kern 1pt} {\kern 1pt} {\kern 1pt} {\kern 1pt} {\kern 1pt} {\kern 1pt} {\kern 1pt} {\kern 1pt} {\kern 1pt} {\kern 1pt} {\kern 1pt} {\kern 1pt}   = {\Theta ^{ *  \top }} - {{\tilde \Theta }^ \top }\left( {n - 1} \right)\\
{\kern 1pt} {\kern 1pt} {\kern 1pt} {\kern 1pt}{\kern 1pt} {\kern 1pt} {\kern 1pt} {\kern 1pt} {\kern 1pt} {\kern 1pt} {\kern 1pt} {\kern 1pt} {\kern 1pt} {\kern 1pt} {\kern 1pt} {\kern 1pt} {\kern 1pt} {\kern 1pt} {\kern 1pt} {\kern 1pt} {\kern 1pt} {\kern 1pt} {\kern 1pt} {\kern 1pt} {\kern 1pt} {\kern 1pt} {\kern 1pt} {\kern 1pt} {\kern 1pt} {\kern 1pt} {\kern 1pt} {\kern 1pt} {\kern 1pt} {\kern 1pt} {\kern 1pt} {\kern 1pt} {\kern 1pt} {\kern 1pt} {\kern 1pt} {\kern 1pt} {\kern 1pt}  + {\bf{H}}\left( n \right){\bf{G}}\left( n \right){{\bf{G}}^ \top }\left( n \right){{\tilde \Theta }^ \top }\left( {n - 1} \right),
\end{array}
\end{equation}
and
\begin{equation}
\label{eq48}
{{\tilde{\Theta }}^{\top }}\left( n \right)=\left( \mathbf{I}-\mathbf{H}\left( n \right)\mathbf{G}\left( n \right){{\mathbf{G}}^{\top }}\left( n \right) \right){{\tilde{\Theta }}^{\top }}\left( n-1 \right),
\end{equation}
where $\mathbf{I}$ is an identity matrix. Substituting Eq. (\ref{eq40}) into Eq. (\ref{eq48}), yields
\begin{equation}
\label{eq49}
{{\tilde{\Theta }}^{\top }}\left( n \right)=\mathbf{H}\left( n \right){{\mathbf{H}}^{-1}}\left( n-1 \right){{\tilde{\Theta }}^{\top }}\left( n-1 \right).
\end{equation}
Let $\mathbf{P}\left( n \right)=\mathbf{H}\left( n \right){{\mathbf{H}}^{-1}}\left( n-1 \right)$, and Eq. (\ref{eq49}) is rewritten as
\begin{equation}
\label{eq50}
{{\tilde{\Theta }}^{\top }}\left( n \right)=\mathbf{P}\left( n \right){{\tilde{\Theta }}^{\top }}\left( n-1 \right).
\end{equation}
Specifically, the matrix $\mathbf{P}\left( n \right)$ can be reconstructed by using Eq. (\ref{eq40})
\begin{equation}
\label{eq51}
\mathbf{P}\left( n \right)={{\left( {{\mathbf{H}}^{-1}}\left( n-1 \right)+\mathbf{G}\left( n \right){{\mathbf{G}}^{\top }}\left( n \right) \right)}^{-1}}{{\mathbf{H}}^{-1}}\left( n-1 \right),
\end{equation}
where ${{\mathbf{H}}^{-1}}\left( n-1 \right)=\sum\limits_{j=1}^{n-1}{\mathbf{G}\left( j \right){{\mathbf{G}}^{\top }}\left( j \right)}$ is a symmetric and positive matrix, and so is $\mathbf{G}\left( n \right){{\mathbf{G}}^{\top }}\left( n \right)$. Therefore, the norm of the matrix $\mathbf{P}\left( n \right)$ is less than 1, that is, $\left\| \mathbf{P}\left( n \right) \right\|<1$. It can be easily inferred that 
\begin{equation}
\label{eq52}
\left\| \tilde{\Theta }\left( n \right) \right\|<\left\| \mathbf{P}\left( n \right) \right\|\left\| \tilde{\Theta }\left( n-1 \right) \right\|.
\end{equation}
Observe that $\underset{n\to \infty }{\mathop{\lim }}\,\tilde{\Theta }\left( n \right)=0$. Then, we obtain $\underset{n\to \infty }{\mathop{\lim }}\,\Theta \left( n \right)={{\Theta }^{*}}$, which verifies the global convergence of the parameters expressed in Eq. (\ref{eq43}) and Eq. (\ref{eq44}).

\section{Performance evaluation}
This section presents some simulation results for evaluating the performance of the proposed F-RSCN model. Three modelling tasks, including a nonlinear system identification and two industrial data predictive analyses, are used in this study. The performance comparisons are conducted among the RSCN \cite{ref13}, fuzzy SCN (F-SCN) \cite{ref5}, fuzzy ESN (F-ESN) \cite{ref14}, and F-RSCN. The normalized root means square error (NRMSE) is employed to evaluate the model performance, that is,
\begin{equation}
\label{eq53}
NRMSE=\sqrt{\frac{\sum\limits_{n=1}^{{{n}_{max}}}{{{\left( \mathbf{y}\left( n \right)-\mathbf{t}\left( n \right) \right)}^{2}}}}{{{n}_{max}}\operatorname{var}\left( \mathbf{t} \right)}},
\end{equation}
where $\operatorname{var}\left( \mathbf{t} \right)$ denotes the variance of the desired output $\mathbf{t}$.

For the parameter settings, following items are taken into account.
\begin{itemize}
  \item  
  The sparsity of the reservoir weight matrix varies from 0.01 to 0.05, and the scaling factor of the spectral radius varies from 0.5 to 1.       
  \item 
  The random parameters for the F-ESN are generated from a uniform distribution $\left[ -1,1 \right]$.  
  \item 
  For RSCNs, the parameters are set as: weight scale sequence $\left\{ 0.1,0.5,1,5,10,50,100\right\}$, contractive sequence $r=\left[ 0.9,0.99,0.999,0.9999 \right]$, the maximum number of stochastic configurations ${{G}_{\max }}=100$, training tolerance ${{\varepsilon }}=0.000001$, and the initial reservoir size is 5.
  \item 
  The number of fuzzy rules and reservoir sizes are determined by the grid search method according to the NRMSE of the validation sets.
  \item 
  Each simulation is performed using 50 independent trials under the same conditions, and the mean and standard deviation of these results are used for comparison.
\end{itemize}

\subsection{Nonlinear system identification}
In this simulation, the problem of nonlinear system identification is considered. Given an unknown dynamic nonlinear plant
\begin{equation}
\label{eq55}
\begin{array}{l}
y\left( {n + 1} \right) = 0.72y\left( n \right) + 0.025y\left( {n - 1} \right)u\left( {n - 1} \right)\\
{\kern 1pt} {\kern 1pt} {\kern 1pt} {\kern 1pt} {\kern 1pt} {\kern 1pt} {\kern 1pt} {\kern 1pt} {\kern 1pt} {\kern 1pt} {\kern 1pt} {\kern 1pt} {\kern 1pt} {\kern 1pt} {\kern 1pt} {\kern 1pt} {\kern 1pt} {\kern 1pt} {\kern 1pt} {\kern 1pt} {\kern 1pt} {\kern 1pt} {\kern 1pt} {\kern 1pt} {\kern 1pt} {\kern 1pt} {\kern 1pt} {\kern 1pt} {\kern 1pt} {\kern 1pt} {\kern 1pt} {\kern 1pt} {\kern 1pt} {\kern 1pt} {\kern 1pt} {\kern 1pt} {\kern 1pt} {\kern 1pt} {\kern 1pt} {\kern 1pt} {\kern 1pt} {\kern 1pt} {\kern 1pt} {\kern 1pt} {\kern 1pt} {\kern 1pt} {\kern 1pt} {\kern 1pt} {\kern 1pt} {\kern 1pt} {\kern 1pt} {\kern 1pt}  + 0.01{u^2}\left( {n - 2} \right) + 0.2u\left( {n - 3} \right),
\end{array}
\end{equation}
in the training phase, the input $u\left( n \right)$ is uniformly generated from $\left[ { - 1,1} \right]$ and the initial output are $y\left( 1 \right)=y\left( 2 \right)=y\left( 3 \right)=0, y\left( 4 \right)=0.1.$ In the testing phase, the input is generated by
\begin{equation}
\label{eq56}
u\left( n \right) = \left\{ {\begin{array}{*{20}{l}}
{\sin \left( {\frac{{\pi n}}{{25}}} \right),{\kern 1pt} {\kern 1pt} {\kern 1pt} {\kern 1pt} {\kern 1pt} {\kern 1pt} {\kern 1pt} {\kern 1pt} {\kern 1pt} {\kern 1pt} {\kern 1pt} {\kern 1pt} {\kern 1pt} {\kern 1pt} {\kern 1pt} {\kern 1pt} {\kern 1pt} {\kern 1pt} {\kern 1pt} {\kern 1pt} {\kern 1pt} {\kern 1pt} {\kern 1pt} {\kern 1pt} {\kern 1pt} {\kern 1pt} {\kern 1pt} {\kern 1pt} {\kern 1pt} {\kern 1pt} {\kern 1pt} {\kern 1pt} {\kern 1pt} {\kern 1pt} {\kern 1pt} {\kern 1pt} {\kern 1pt} {\kern 1pt} {\kern 1pt} {\kern 1pt} {\kern 1pt} {\kern 1pt} {\kern 1pt} {\kern 1pt} {\kern 1pt} {\kern 1pt} {\kern 1pt} {\kern 1pt} {\kern 1pt} {\kern 1pt} {\kern 1pt} {\kern 1pt} {\kern 1pt} {\kern 1pt} {\kern 1pt} {\kern 1pt} {\kern 1pt} {\kern 1pt} {\kern 1pt} {\kern 1pt} {\kern 1pt} {\kern 1pt} {\kern 1pt} {\kern 1pt} {\kern 1pt} {\kern 1pt} {\kern 1pt} {\kern 1pt} {\kern 1pt} {\kern 1pt} {\kern 1pt} {\kern 1pt} {\kern 1pt} {\kern 1pt} {\kern 1pt} {\kern 1pt} {\kern 1pt} {\kern 1pt} {\kern 1pt} {\kern 1pt} {\kern 1pt} {\kern 1pt} {\kern 1pt} {\kern 1pt} {\kern 1pt} {\kern 1pt} {\kern 1pt} {\kern 1pt} {\kern 1pt} {\kern 1pt} {\kern 1pt} {\kern 1pt} {\kern 1pt} {\kern 1pt} {\kern 1pt} {\kern 1pt} {\kern 1pt} {\kern 1pt} {\kern 1pt} {\kern 1pt} {\kern 1pt} {\kern 1pt} 0 < n < 250}\\
{1,{\kern 1pt}  {\kern 1pt} {\kern 1pt} {\kern 1pt} {\kern 1pt} {\kern 1pt} {\kern 1pt} {\kern 1pt} {\kern 1pt} {\kern 1pt} {\kern 1pt} {\kern 1pt} {\kern 1pt} {\kern 1pt} {\kern 1pt} {\kern 1pt} {\kern 1pt} {\kern 1pt} {\kern 1pt} {\kern 1pt} {\kern 1pt} {\kern 1pt} {\kern 1pt} {\kern 1pt} {\kern 1pt} {\kern 1pt} {\kern 1pt} {\kern 1pt} {\kern 1pt} {\kern 1pt} {\kern 1pt} {\kern 1pt} {\kern 1pt} {\kern 1pt} {\kern 1pt} {\kern 1pt} {\kern 1pt} {\kern 1pt} {\kern 1pt} {\kern 1pt} {\kern 1pt} {\kern 1pt} {\kern 1pt} {\kern 1pt} {\kern 1pt} {\kern 1pt} {\kern 1pt} {\kern 1pt} {\kern 1pt} {\kern 1pt} {\kern 1pt} {\kern 1pt} {\kern 1pt} {\kern 1pt} {\kern 1pt} {\kern 1pt}{\kern 1pt} {\kern 1pt} {\kern 1pt} {\kern 1pt} {\kern 1pt} {\kern 1pt} {\kern 1pt} {\kern 1pt} {\kern 1pt} {\kern 1pt} {\kern 1pt} {\kern 1pt} {\kern 1pt} {\kern 1pt} {\kern 1pt} {\kern 1pt} {\kern 1pt} {\kern 1pt} {\kern 1pt} {\kern 1pt} {\kern 1pt} {\kern 1pt} {\kern 1pt} {\kern 1pt} {\kern 1pt} {\kern 1pt} {\kern 1pt} {\kern 1pt} {\kern 1pt} {\kern 1pt} {\kern 1pt} {\kern 1pt} {\kern 1pt} {\kern 1pt} {\kern 1pt} {\kern 1pt} {\kern 1pt} {\kern 1pt} {\kern 1pt} {\kern 1pt} {\kern 1pt} {\kern 1pt} {\kern 1pt} {\kern 1pt} {\kern 1pt} {\kern 1pt} {\kern 1pt} {\kern 1pt} {\kern 1pt} {\kern 1pt} {\kern 1pt} {\kern 1pt} {\kern 1pt} {\kern 1pt} {\kern 1pt} {\kern 1pt} {\kern 1pt} {\kern 1pt} {\kern 1pt} {\kern 1pt} {\kern 1pt} {\kern 1pt} {\kern 1pt} {\kern 1pt} {\kern 1pt} {\kern 1pt} {\kern 1pt} {\kern 1pt} {\kern 1pt} {\kern 1pt} 250 \le n < 500}\\
{ - 1,{\kern 1pt}  {\kern 1pt} {\kern 1pt}  {\kern 1pt} {\kern 1pt} {\kern 1pt} {\kern 1pt} {\kern 1pt} {\kern 1pt} {\kern 1pt}{\kern 1pt} {\kern 1pt} {\kern 1pt} {\kern 1pt} {\kern 1pt} {\kern 1pt} {\kern 1pt} {\kern 1pt} {\kern 1pt} {\kern 1pt} {\kern 1pt} {\kern 1pt} {\kern 1pt} {\kern 1pt} {\kern 1pt} {\kern 1pt} {\kern 1pt} {\kern 1pt} {\kern 1pt} {\kern 1pt} {\kern 1pt} {\kern 1pt} {\kern 1pt} {\kern 1pt} {\kern 1pt} {\kern 1pt} {\kern 1pt} {\kern 1pt} {\kern 1pt} {\kern 1pt} {\kern 1pt} {\kern 1pt} {\kern 1pt} {\kern 1pt} {\kern 1pt} {\kern 1pt} {\kern 1pt} {\kern 1pt} {\kern 1pt} {\kern 1pt} {\kern 1pt} {\kern 1pt} {\kern 1pt} {\kern 1pt} {\kern 1pt} {\kern 1pt} {\kern 1pt} {\kern 1pt} {\kern 1pt} {\kern 1pt} {\kern 1pt} {\kern 1pt} {\kern 1pt} {\kern 1pt} {\kern 1pt} {\kern 1pt} {\kern 1pt} {\kern 1pt} {\kern 1pt} {\kern 1pt} {\kern 1pt} {\kern 1pt} {\kern 1pt} {\kern 1pt} {\kern 1pt} {\kern 1pt} {\kern 1pt} {\kern 1pt} {\kern 1pt} {\kern 1pt} {\kern 1pt} {\kern 1pt} {\kern 1pt} {\kern 1pt} {\kern 1pt} {\kern 1pt} {\kern 1pt} {\kern 1pt} {\kern 1pt} {\kern 1pt} {\kern 1pt} {\kern 1pt} {\kern 1pt} {\kern 1pt} {\kern 1pt} {\kern 1pt} {\kern 1pt} {\kern 1pt} {\kern 1pt} {\kern 1pt} {\kern 1pt} {\kern 1pt} {\kern 1pt} {\kern 1pt} {\kern 1pt} {\kern 1pt} {\kern 1pt} {\kern 1pt} {\kern 1pt} {\kern 1pt} {\kern 1pt} {\kern 1pt} {\kern 1pt} {\kern 1pt} {\kern 1pt} {\kern 1pt} {\kern 1pt} {\kern 1pt} 500 \le n < 750}\\
\begin{array}{l}
0.6\cos \left( {\frac{{\pi n}}{{10}}} \right) + 0.1\cos \left( {\frac{{\pi n}}{{32}}} \right) + \\
{\kern 1pt} {\kern 1pt} {\kern 1pt} {\kern 1pt} {\kern 1pt} 0.3\sin \left( {\frac{{\pi n}}{{25}}} \right),{\kern 1pt} {\kern 1pt} {\kern 1pt} {\kern 1pt} {\kern 1pt} {\kern 1pt} {\kern 1pt} {\kern 1pt} {\kern 1pt} {\kern 1pt} {\kern 1pt} {\kern 1pt} {\kern 1pt}  {\kern 1pt} {\kern 1pt} {\kern 1pt} {\kern 1pt} {\kern 1pt} {\kern 1pt} {\kern 1pt} {\kern 1pt} {\kern 1pt} {\kern 1pt} {\kern 1pt} {\kern 1pt} {\kern 1pt} {\kern 1pt} {\kern 1pt} {\kern 1pt} {\kern 1pt} {\kern 1pt} {\kern 1pt} {\kern 1pt} {\kern 1pt} {\kern 1pt} {\kern 1pt} {\kern 1pt} {\kern 1pt} {\kern 1pt} {\kern 1pt} {\kern 1pt} {\kern 1pt} {\kern 1pt} {\kern 1pt} {\kern 1pt} {\kern 1pt} {\kern 1pt} {\kern 1pt} {\kern 1pt} {\kern 1pt} {\kern 1pt} {\kern 1pt} {\kern 1pt} {\kern 1pt} {\kern 1pt} {\kern 1pt} {\kern 1pt} {\kern 1pt} {\kern 1pt} {\kern 1pt} {\kern 1pt} {\kern 1pt} {\kern 1pt} {\kern 1pt} {\kern 1pt} {\kern 1pt} {\kern 1pt} 750 \le n \le 1000.
\end{array}
\end{array}} \right.
\end{equation}
A total of 2000, 1000, and 1000 samples are generated as the training, validation, and testing set, respectively. The first 100 samples of each set are washed. Specifically, considering the order uncertainty, $\left[ y\left( n \right),u\left( n \right) \right]$ is used to predict $y\left( n+1 \right)$.

\begin{figure}[htpb]
	\centering
	\subfloat[Prediction results]{\includegraphics[width=8cm]{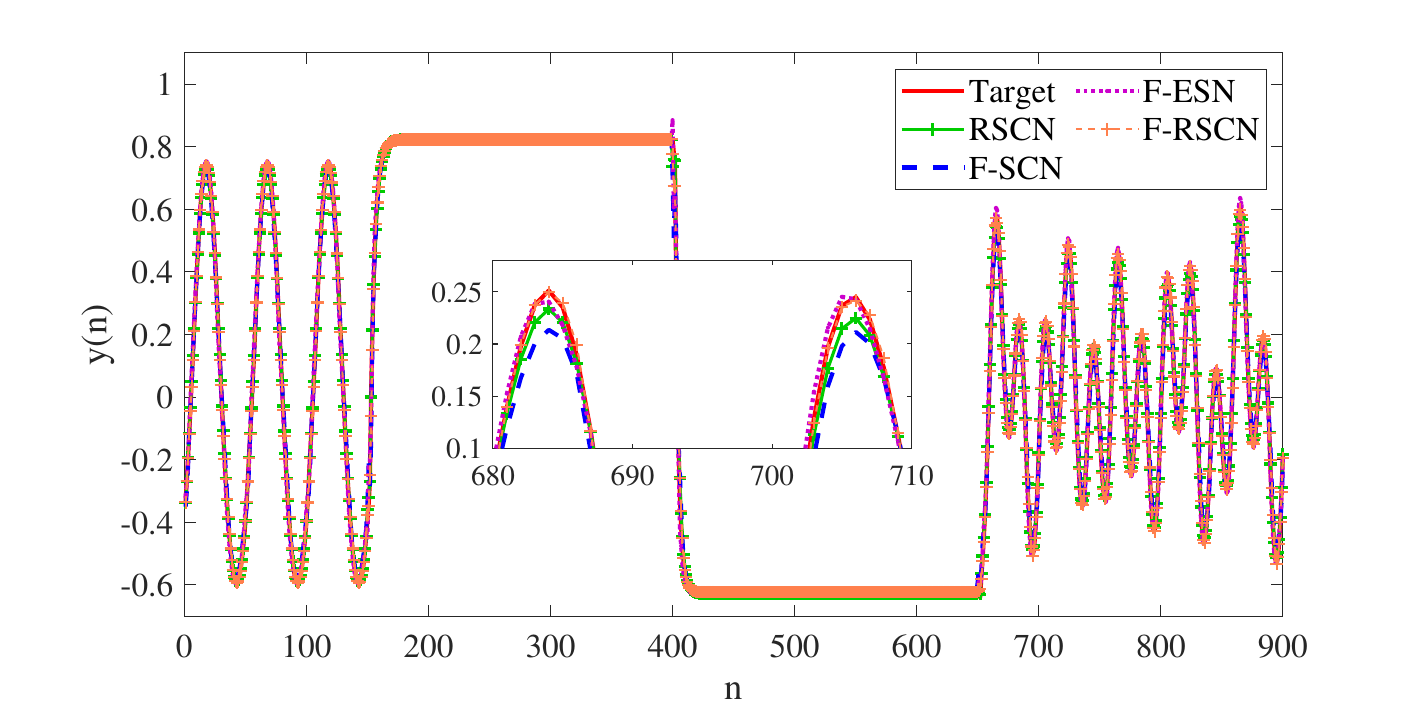}}\\
	\subfloat[Prediction error values]{\includegraphics[width=8cm]{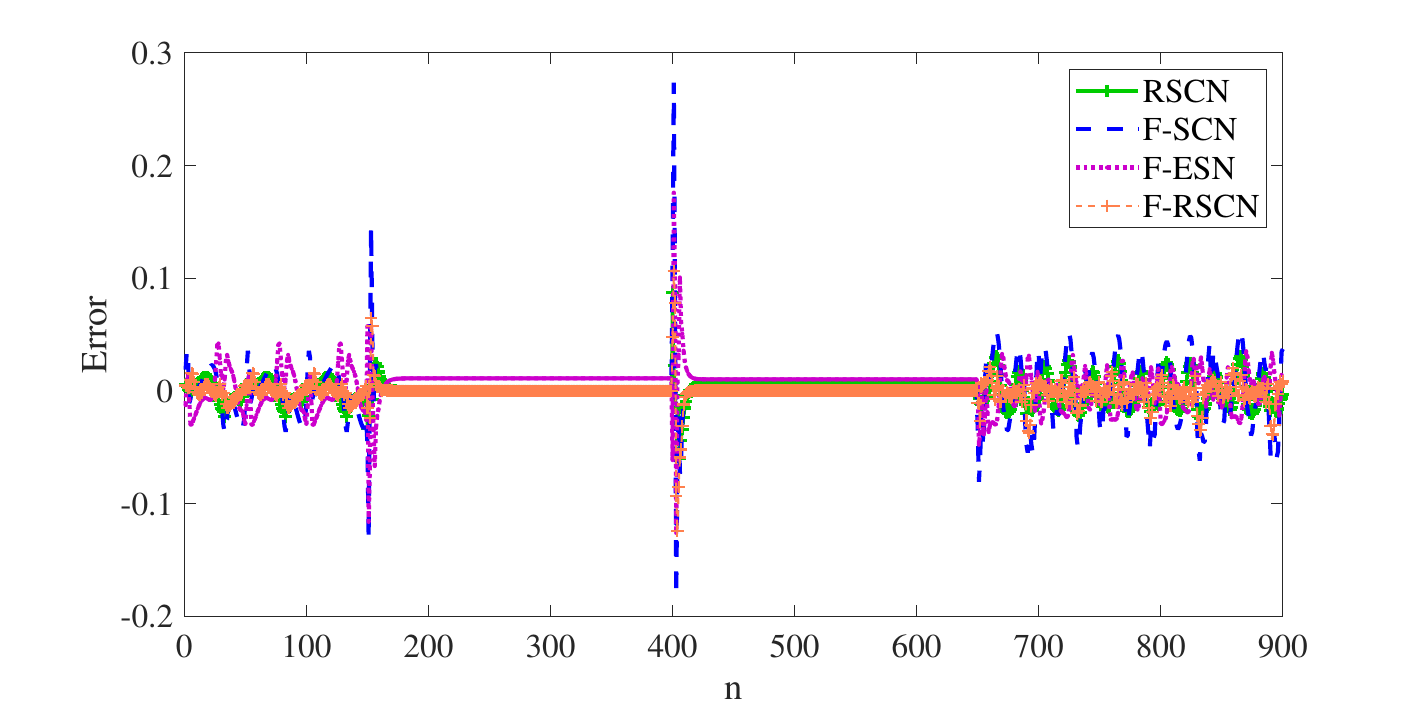}}
	\caption{Prediction fitting curves and error values of different models for nonlinear system identification task.}
	\label{fig2}
\end{figure}

\begin{figure}[htpb]
	\centering
	\subfloat{\includegraphics[width=2.6in]{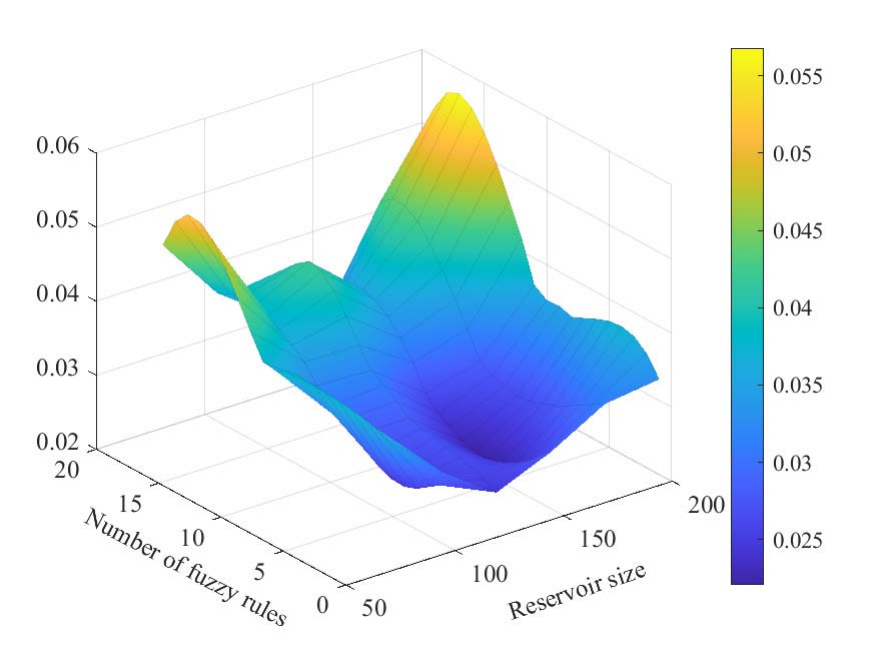}}
	\caption{The testing NRMSE surface map of the F-RSCN with different number of fuzzy rules and reservoir sizes on the nonlinear system identification task.}
	\label{fig3}
\end{figure}

Fig.~\ref{fig2} exhibits the prediction fitting curves and error values generated by the RSCN, F-SCN, F-ESN, and F-RSCN for the nonlinear system identification task. It is evident that F-RSCNs outperform other models in fitting desired outputs, with the smallest error ranges, thus confirming the effectiveness of the proposed method. To investigate the impact of the number of fuzzy rules \emph{Q} and reservoir size \emph{N} on our proposed F-RSCNs, we assess model performance using various combinations of \emph{N} and \emph{Q}. As depicted in Fig.~\ref{fig3}, the testing NRMSE fluctuates with different \emph{N} and \emph{Q}. Excessively large reservoir sizes and an abundance of fuzzy rules lead to subpar prediction results, underscoring the significance of finding an optimal combination of \emph{N} and \emph{Q} for achieving good generalization performance.

\begin{table}[htpb]
\setlength\tabcolsep{3pt}
\centering
  \caption{Performance comparison of different models on the nonlinear system identification task\label{tab1}}
\begin{tabular}{ccccc}
\hline
\multirow{2}{*}{Models} & \multicolumn{4}{c}{Training-testing NRMSE with different number of fuzzy rules} \\ \cline{2-5} 
                        & \textit{Q}=3                      & \textit{Q}=5                      & \textit{Q}=10                     & \textit{Q}=15                     \\ \hline
RSCN                    & 0.0079-0.0395          & -                        & -                        & -                        \\
F-SCN                   & 0.0166-0.0567          & 0.0122-0.0459          & 0.0106-0.0398          & 0.0092-0.0384          \\
F-ESN                   & 0.0094-0.0464          & 0.0084-0.0423          & 0.0080-0.0422          & 0.0085-0.0433          \\
F-RSCN                  & \textbf{0.0039-0.0268}          & \textbf{0.0038-0.0253}          & \textbf{0.0035-0.0263}          & \textbf{0.0036-0.0258}          \\ \hline
\end{tabular}
\end{table}

\begin{figure}[htpb]
\centering
	\subfloat{\includegraphics[width=3.5in]{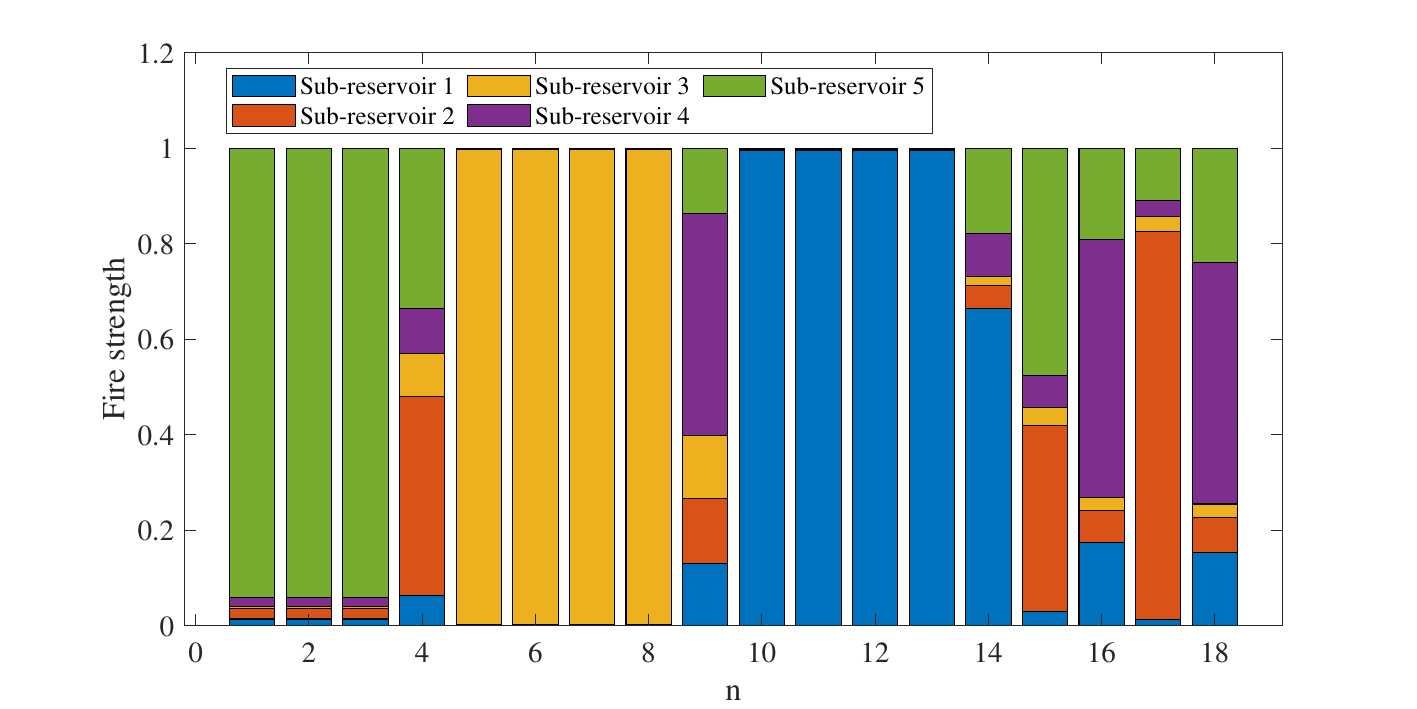}}
\caption{The fire strength of the F-RSCN for different testing samples on the nonlinear system identification task (\textit{Q}=5).}
\label{fig4}
\end{figure}

The performance comparisons of the nonlinear system identification tasks are detailed in Table~\ref{tab1}. The proposed F-RSCNs demonstrate sound performance on both the training and testing sets, further validating the effectiveness of F-RSCNs in nonlinear system identification. Increasing the number of fuzzy rules can improve the model's complexity and accuracy, which is essential for system identification tasks that require high precision. The performance of F-ESN and F-SCN tends to improve with a larger \emph{Q}, suggesting less stability compared to F-RSCN. F-RSCN consistently demonstrates optimal performance across different numbers of fuzzy rules, underscoring its robustness to changes in fuzzy variables. 

Fig.~\ref{fig4} depicts the fire strength of the F-RSCN for various testing samples in the nonlinear system identification task, with sampling frequencies of 50. It is observed that sub-reservoir 5 demonstrates higher fire strength in the initial operation stages, suggesting a more significant contribution to the model output during the early phases, possibly due to its effective capture of the system's initial state. However, as the system continues to operate, the fire strength of sub-reservoir 3 and sub-reservoir 1 gradually increases, indicating an increasing importance of these sub-models in influencing the system's dynamic behavior in later stages. The incorporation of the fuzzy inference module allows the model to assign appropriate weights to each sub-model by establishing rational fuzzy rules based on input information. This not only enhances the model's flexibility but also improves its ability to adapt to the system's dynamic behavior. By monitoring these fluctuations, a better understanding and explanation of the system's behavioral characteristics at different operational stages can be achieved, leading to enhanced prediction accuracy and interpretability of the model.

\subsection{Soft sensing of butane in dehumanizer column}
\begin{figure}[htpb]
\centering
\includegraphics[width=2.2in]{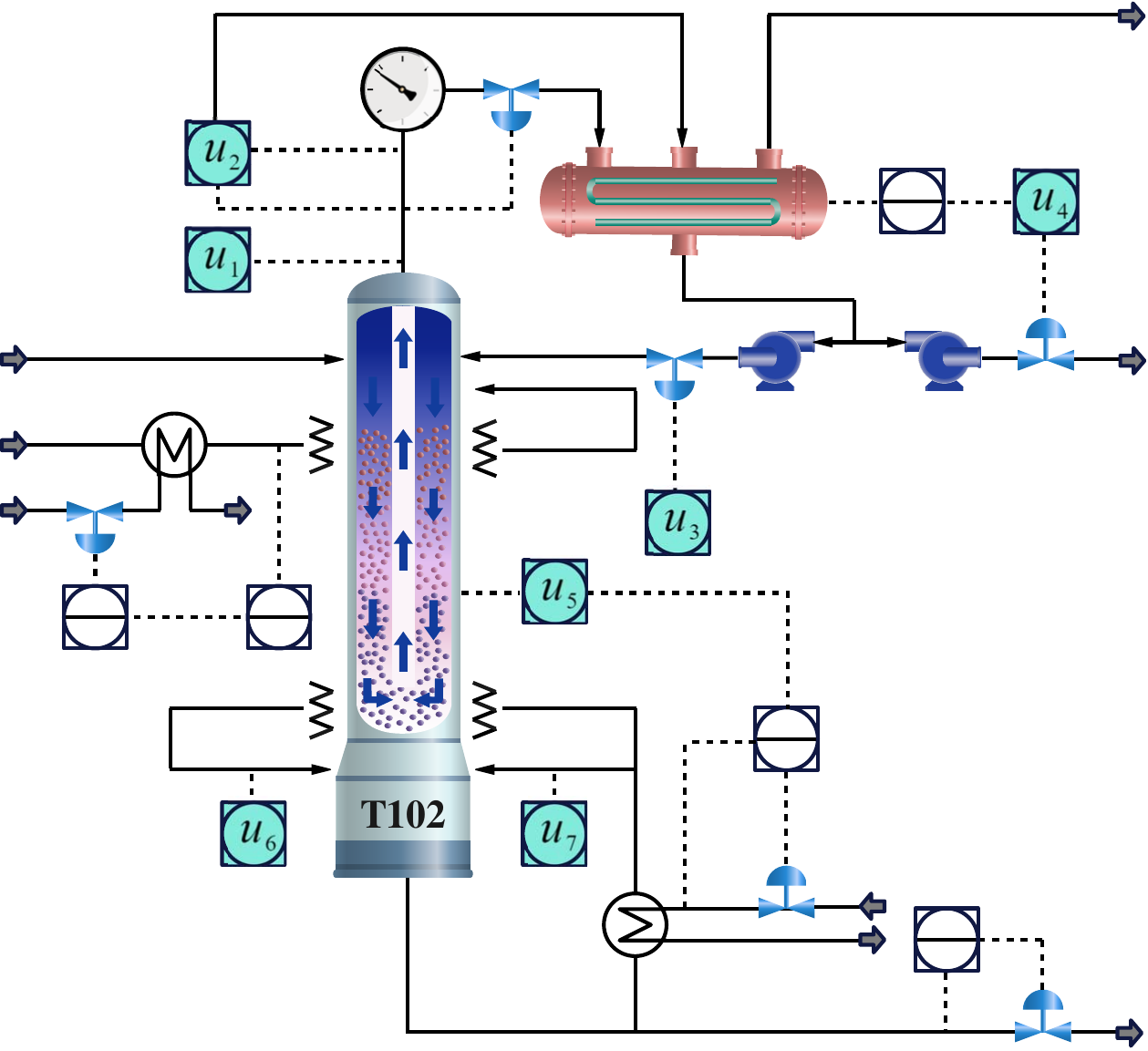}
\caption{Flowchart of debutanizer column.}
\label{fig5}
\end{figure}

The debutanizer column is a key component in the naphtha fractionation process, specifically in separating butane (C4) from naphtha. Real-time monitoring of butane concentration is vital for process optimization and reducing butane content at the tower's base. However, directly measuring butane concentration can be difficult and may have a delay. Therefore, it is important to develop a precise soft sensing model. The specific operation flow of the dehumanizer column is shown in Fig.~\ref{fig5}, and the relevant auxiliary variables include the tower top temperature ${{u}_{1}}$, tower top pressure ${{u}_{2}}$, tower top reflux flow ${{u}_{3}}$, tower top product outflow ${{u}_{4}}$, 6-th tray temperature ${{u}_{5}}$, tower bottom temperature ${{u}_{6}}$, and tower bottom pressure ${{u}_{7}}$. To conveniently analyze and model, Fortuna et al. \cite{ref34} presented a well-designed combination of variables to obtain the butane concentration $y\left( n \right)$,
\begin{equation}
\label{eq59}
\begin{array}{l}
y\left( n \right) = f\left( {{u_1}\left( n \right),} \right.{u_2}\left( n \right),{u_3}\left( n \right),{u_4}\left( n \right),{u_5}\left( n \right),{u_5}\left( {n - 1} \right),\\
{\kern 1pt} {\kern 1pt} {\kern 1pt} {\kern 1pt} {\kern 1pt} {\kern 1pt} {\kern 1pt} {\kern 1pt} {\kern 1pt} {\kern 1pt} {\kern 1pt} {\kern 1pt} {\kern 1pt} {\kern 1pt} {\kern 1pt} {\kern 1pt} {\kern 1pt} {\kern 1pt} {\kern 1pt} {\kern 1pt} {\kern 1pt} {\kern 1pt} {\kern 1pt} {\kern 1pt} {\kern 1pt} {\kern 1pt} {\kern 1pt} {\kern 1pt} {\kern 1pt} {\kern 1pt} {\kern 1pt} {\kern 1pt} {\kern 1pt} {\kern 1pt} {\kern 1pt} {\kern 1pt} {\kern 1pt} {\kern 1pt} {\kern 1pt} {\kern 1pt} {\kern 1pt} {\kern 1pt} {\kern 1pt} {\kern 1pt} {\kern 1pt} {\kern 1pt} {\kern 1pt} {\kern 1pt} {\kern 1pt} {\kern 1pt} {u_5}\left( {n - 2} \right),{u_5}\left( {n - 3} \right),\left( {{u_1}\left( n \right) + {u_2}\left( n \right)} \right)/2,\\
{\kern 1pt} {\kern 1pt} {\kern 1pt} {\kern 1pt} {\kern 1pt} {\kern 1pt} {\kern 1pt} {\kern 1pt} {\kern 1pt} {\kern 1pt} {\kern 1pt} {\kern 1pt} {\kern 1pt} {\kern 1pt} {\kern 1pt} {\kern 1pt} {\kern 1pt} {\kern 1pt} {\kern 1pt} {\kern 1pt} {\kern 1pt} {\kern 1pt} {\kern 1pt} {\kern 1pt} {\kern 1pt} {\kern 1pt} {\kern 1pt} {\kern 1pt} {\kern 1pt} {\kern 1pt} {\kern 1pt} {\kern 1pt} {\kern 1pt} {\kern 1pt} {\kern 1pt} {\kern 1pt} {\kern 1pt} {\kern 1pt} {\kern 1pt} {\kern 1pt} {\kern 1pt} {\kern 1pt} {\kern 1pt} {\kern 1pt} {\kern 1pt} {\kern 1pt} {\kern 1pt} \left. {{\kern 1pt} {\kern 1pt} {\kern 1pt} y\left( {n - 1} \right),y\left( {n - 2} \right),y\left( {n - 3} \right),y\left( {n - 4} \right)} \right).
\end{array}
\end{equation}
In our simulation, considering the order uncertainty, $y\left( n \right)$ is predicted by $\left[ {{u}_{1}}\left( n \right), \right.{{u}_{2}}\left( n \right),{{u}_{3}}\left( n \right),{{u}_{4}}\left( n \right),$ $\left. {{u}_{5}}\left( n \right),y\left( n-1 \right) \right]$. A total of 2394 samples are generated. The first 1500 samples from $1\le n\le 1500$ are chosen as the training set. The last 894 samples from $1501\le n\le 2394$ are chosen as the testing set and Gaussian white noise is added to the testing set to generate the validation set. The first 100 samples of each set are washed out.
\begin{figure}[htpb]
\centering
\includegraphics[width=2.8in]{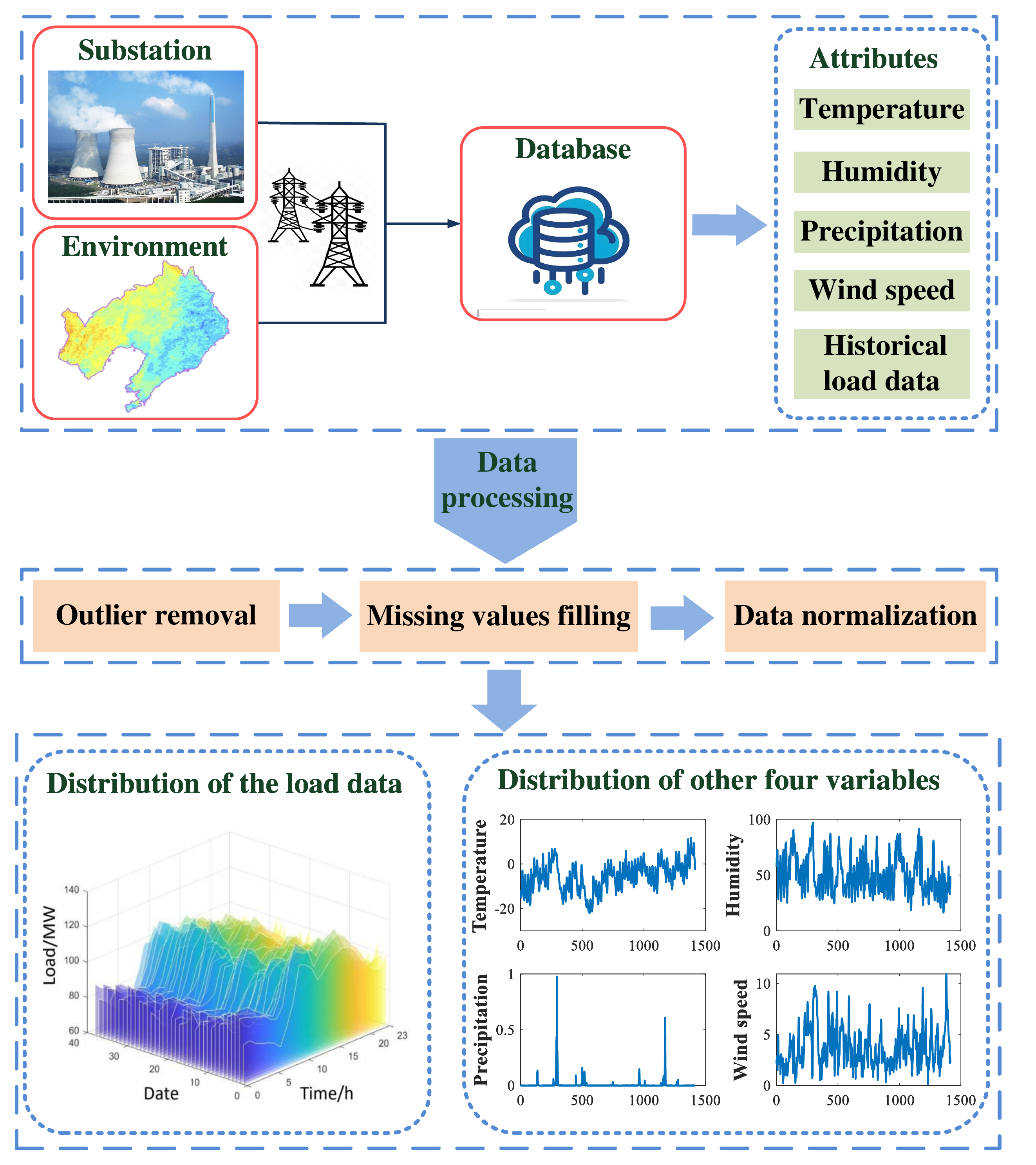}
\caption{Flowchart of the data collection and processing for the short-term power load forecast.}
\label{fig6}
\end{figure}
\begin{figure}[htpb]
	\centering
	\subfloat[Prediction results]{\includegraphics[width=8cm]{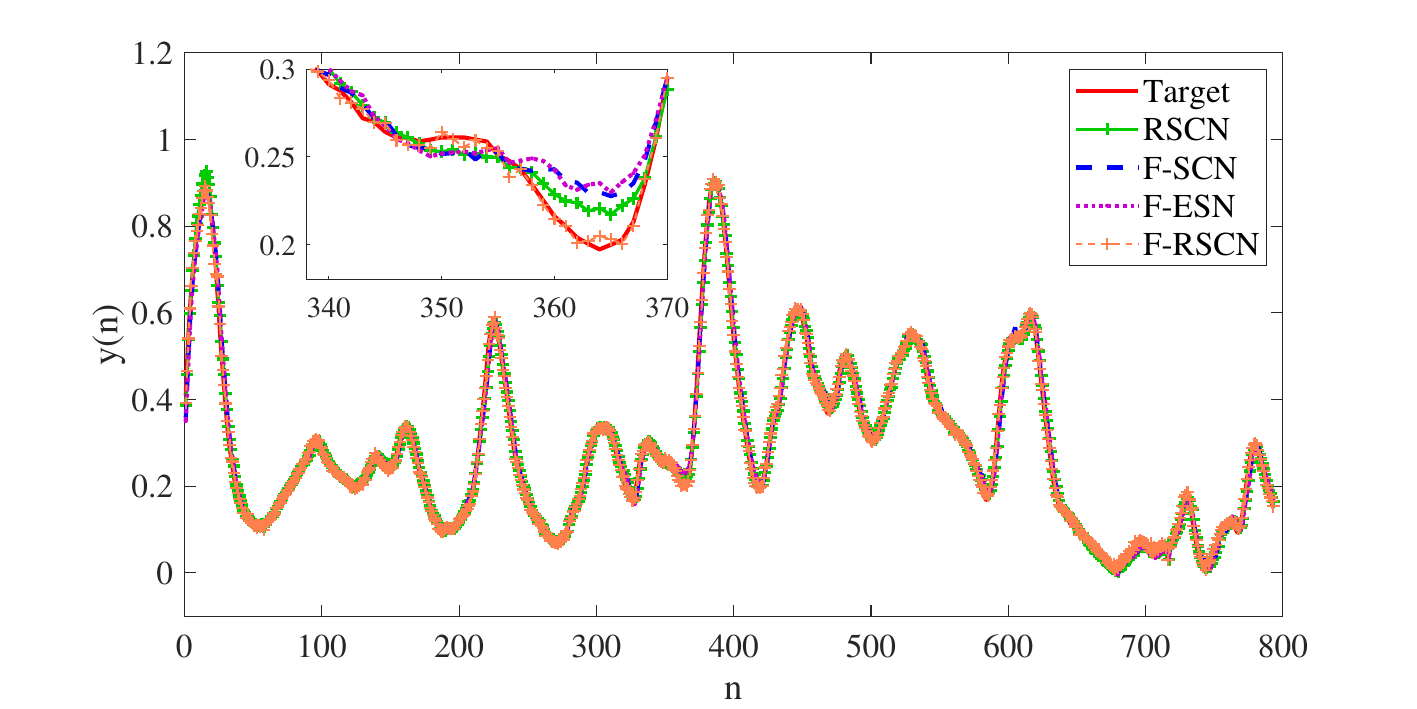}}\\
	\subfloat[Prediction error values]{\includegraphics[width=8cm]{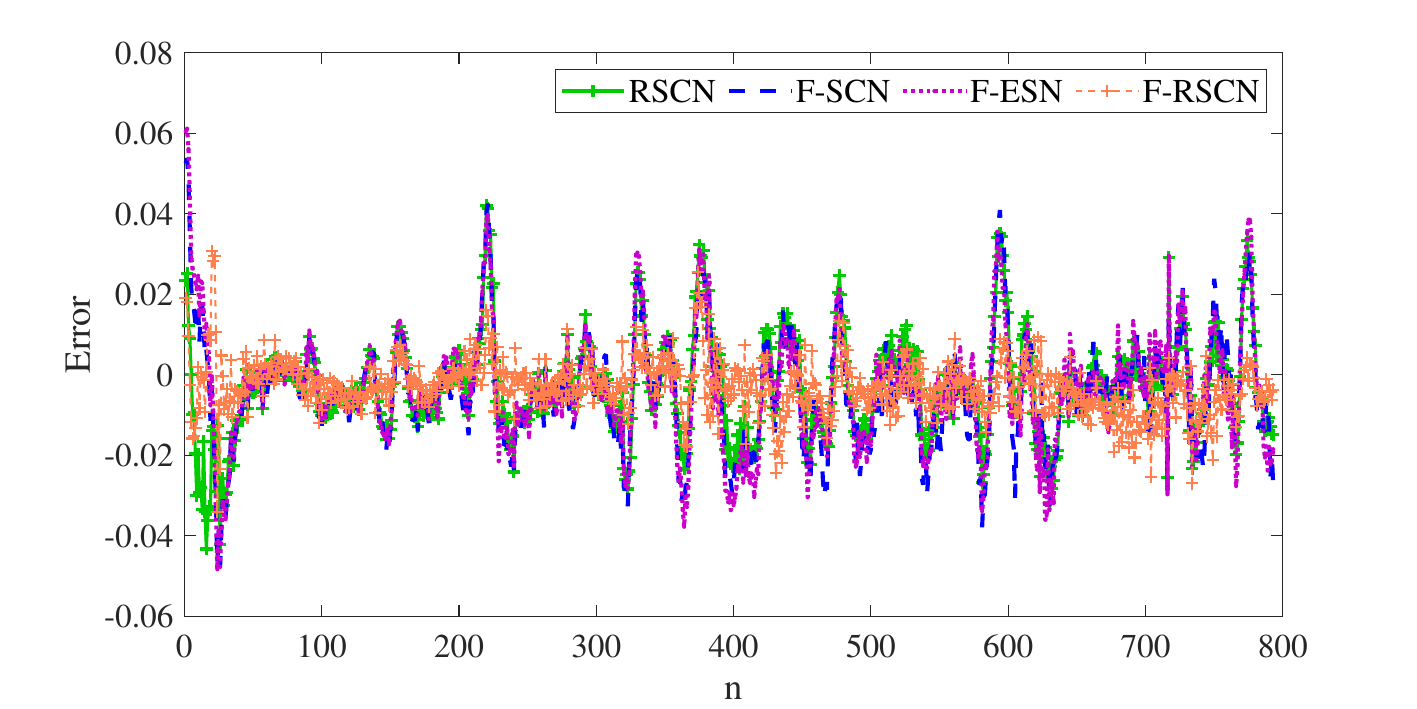}}
	\caption{Prediction fitting curves and error values of different models for the debutanizer column.}
	\label{fig7}
\end{figure}

\subsection{Short-term power load forecasting}
Short-term electricity load forecasting is essential for the reliable and cost-effective operation of power systems. This study utilizes load data from a 500kV substation in Liaoning Province, China, collected hourly over a period of 59 days from January to February 2023. Environmental factors such as temperature ${u_1}$, humidity ${u_2}$, precipitation ${u_3}$, and wind speed ${u_4}$ are considered in predicting the electricity load $y$. The process of data collection and analysis for short-term electricity load forecasting is illustrated in Fig.~\ref{fig6}. The dataset consists of 1415 samples, with 1000 for training and 415 for testing. Gaussian noise is introduced to the test set to form a validation set. To address the order uncertainty, $\left[ {{u}_{1}}\left( n \right), \right.$$\left. {{u}_{2}}\left( n \right),{{u}_{3}}\left( n \right),{{u}_{4}}\left( n \right),y\left( n-1 \right) \right]$ is utilized for forecasting $y\left( n \right)$ in the experiment, with the initial 30 samples of each set excluded.

\begin{figure}[htpb]
	\centering
	\subfloat[Prediction results]{\includegraphics[width=8cm]{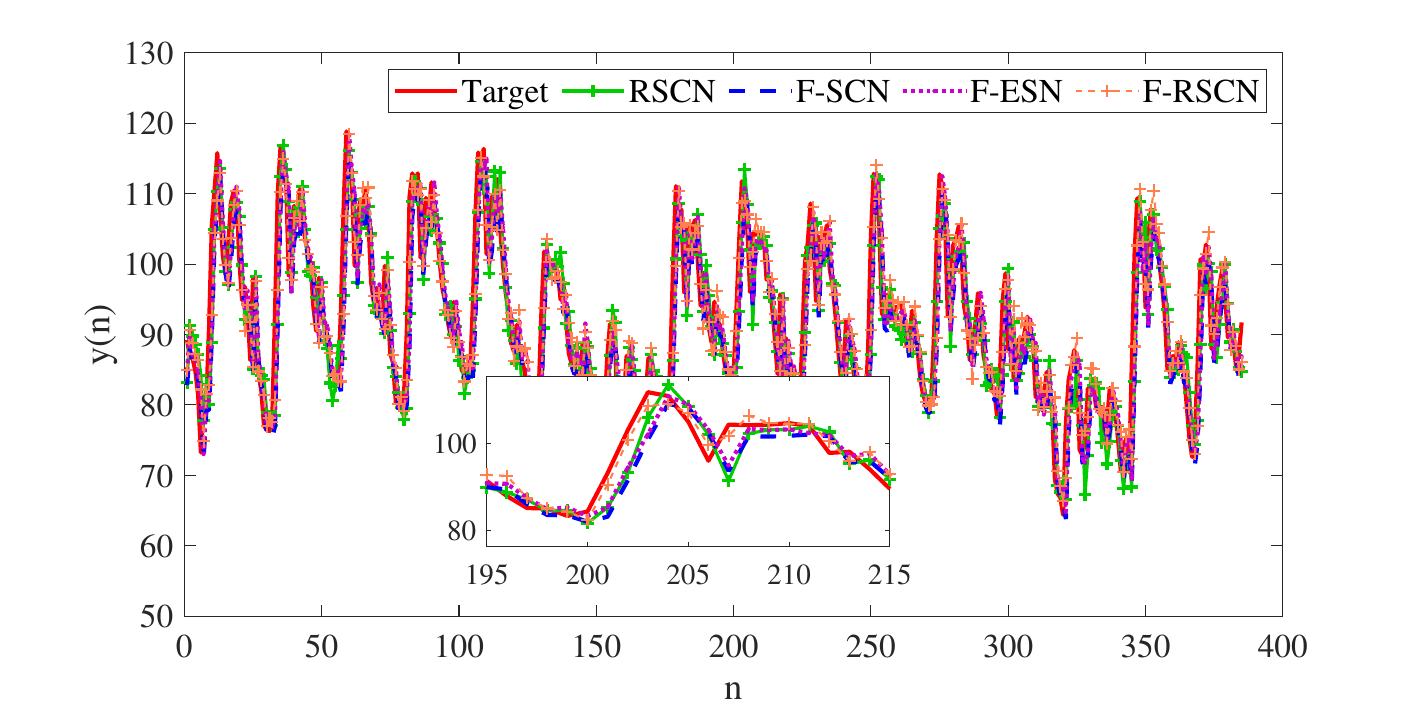}}\\
	\subfloat[Prediction error values]{\includegraphics[width=8cm]{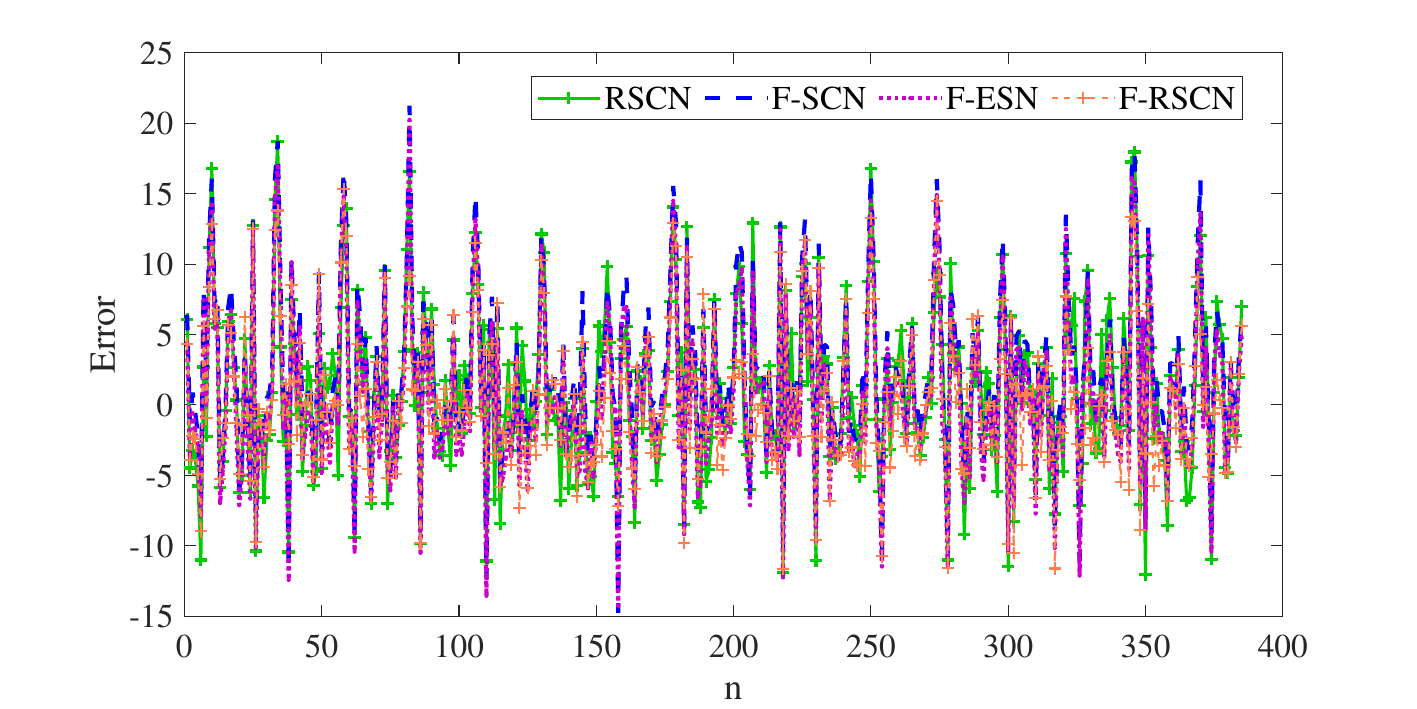}}
	\caption{Prediction fitting curves and error values of different models for the short-term power load.}
	\label{fig8}
\end{figure}

\begin{figure}[htpb]
\centering
	\subfloat[Case 1]{\includegraphics[width=2.6in]{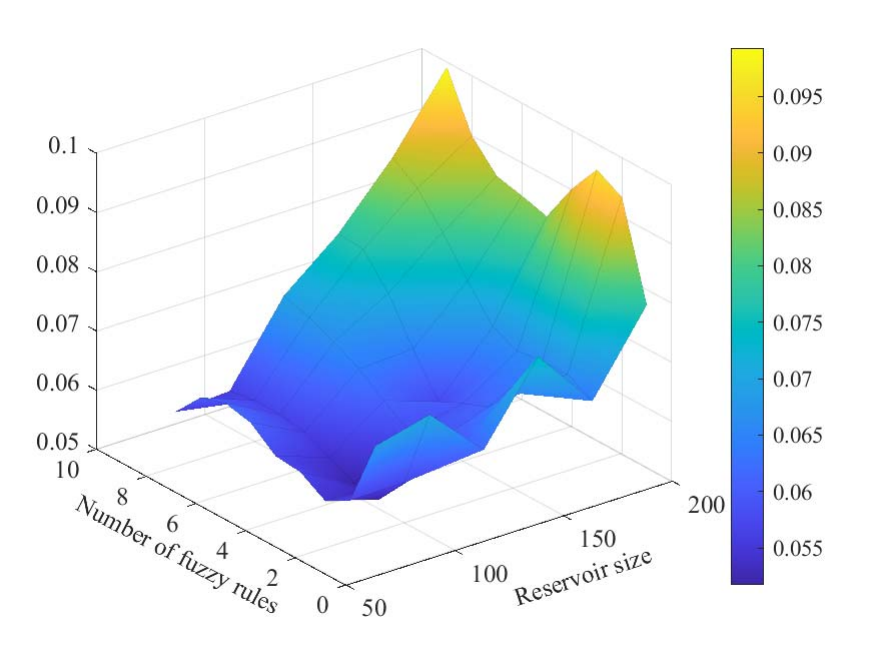}}\\
	\subfloat[Case 2]{\includegraphics[width=2.6in]{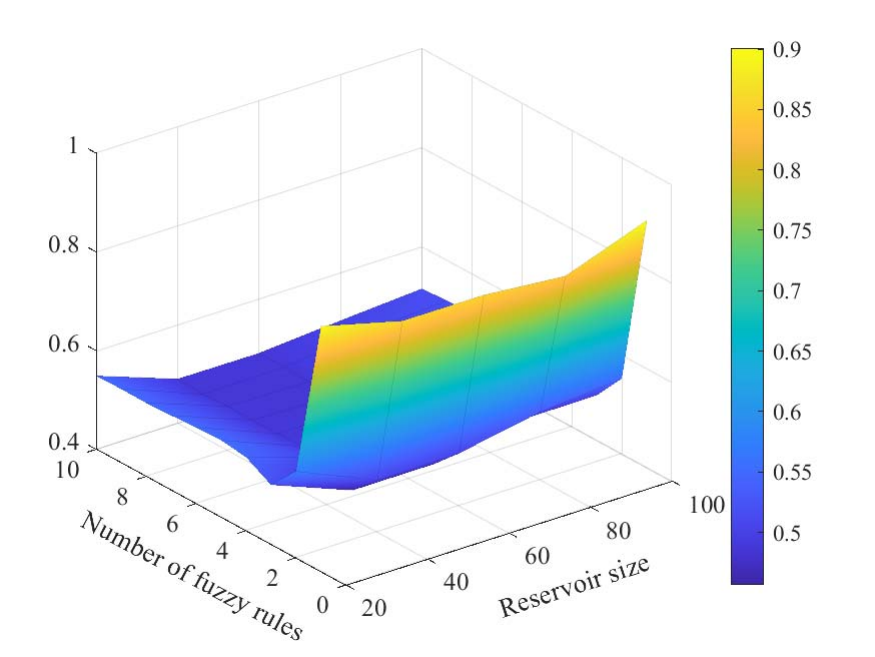}}
\caption{The testing NRMSE surface map of the F-RSCN with different number of fuzzy rules and reservoir sizes on the two industry cases.}
\label{fig9}
\end{figure}

To demonstrate the validity of our proposed scheme, we compare the F-RSCN with the RSCN, F-SCN, and F-ESN, and the prediction results and errors of each model are shown in Fig.~\ref{fig7} and Fig.~\ref{fig8}. The results clearly demonstrate that F-RSCNs outperform the other models, achieving a smaller prediction error range of $\left[ -0.033,0.027 \right]$ and $\left[ -11.66,15.32 \right]$ for the two industry cases, respectively. The testing performance of the proposed F-RSCN with different combinations of fuzzy rules \emph{Q} and reservoir size \emph{N} is plotted in Fig.~\ref{fig9}. It can be seen that the testing NRMSE is sensitive to both parameters, and the best prediction result is obtained when \emph{Q}=4, \emph{N}=100 and \emph{Q}=3, \emph{N}=50 for the two industry tasks.
To intuitively compare the modelling performance of the proposed F-RSCN with other models, we summarize the experimental results in Table~\ref{tab2}. It can be found that F-RSCNs exhibit smaller NRMSE in both training and testing datasets. Furthermore, they consistently outperform other models across varying numbers of fuzzy rules. Although F-SCNs and F-ESNs also show improved performance with an increase in fuzzy rules, they do not exhibit the same level of stability as F-RSCNs. Notably, F-SCNs display a substantial performance boost in industrial case 2 but a more modest improvement in industrial case 1, indicating some level of instability. The stability and reliability of F-RSCNs with different numbers of fuzzy rules make them an optimal solution for addressing nonlinear and uncertain challenges in industrial processes. When compared to alternative models, F-RSCNs excel in capturing and explaining the dynamic behavior of complex systems, providing more accurate and reliable predictions. These superior performance and robust characteristics highlight their potential applications in modelling industrial data.
\begin{table}[htbp]
\scriptsize
\setlength\tabcolsep{3pt}
\centering
  \caption{Performance comparison of different models on the two industry cases\label{tab2}}
\begin{tabular}{cccccc}
\hline
\multirow{2}{*}{Datasets} & \multirow{2}{*}{Models} & \multicolumn{4}{c}{Training-testing NRMSE with different number of fuzzy rules} \\ \cline{3-6} 
                          &                         & \textit{Q}=3                      & \textit{Q}=5                      & \textit{Q}=10                     & \textit{Q}=15                     \\ \hline
\multirow{4}{*}{Case 1}   & RSCN                    & \textbf{0.0273}-0.0679          & -                        & -                        & -                        \\
                          & F-SCN                   & 0.0317-0.0710          & 0.0308-0.0714          & 0.0293-0.0674          & 0.0292-0.0661          \\
                          & F-ESN                   & 0.0272-0.0679          & 0.0277-0.0673          & 0.0273-0.0665          & 0.0277-0.0664          \\
                          & F-RSCN                  & 0.0275-\textbf{0.0556} & \textbf{0.0267-0.0529} & \textbf{0.0265-0.0535} & \textbf{0.0275-0.0531} \\ \hline
\multirow{4}{*}{Case 2}   & RSCN                    & 0.1925-0.5107          & -                        & -                        & -                        \\
                          & F-SCN                   & 0.2503-0.7189          & 0.2495-0.6937         & 0.2387-0.6522         & 0.2314-0.6239         \\
                          & F-ESN                   & 0.2265-0.6338          & 0.2119-0.6267          & 0.2133-0.6236         & 0.2189-0.6324         \\
                          & F-RSCN                  & \textbf{0.1920-0.5101} & \textbf{0.1925-0.4935} & \textbf{0.1934-0.4967} & \textbf{0.1911-0.5007} \\ \hline
\end{tabular}
\end{table}

\begin{figure*}[htpb]
\centering
	\subfloat[Case 1]{\includegraphics[width=3.5in]{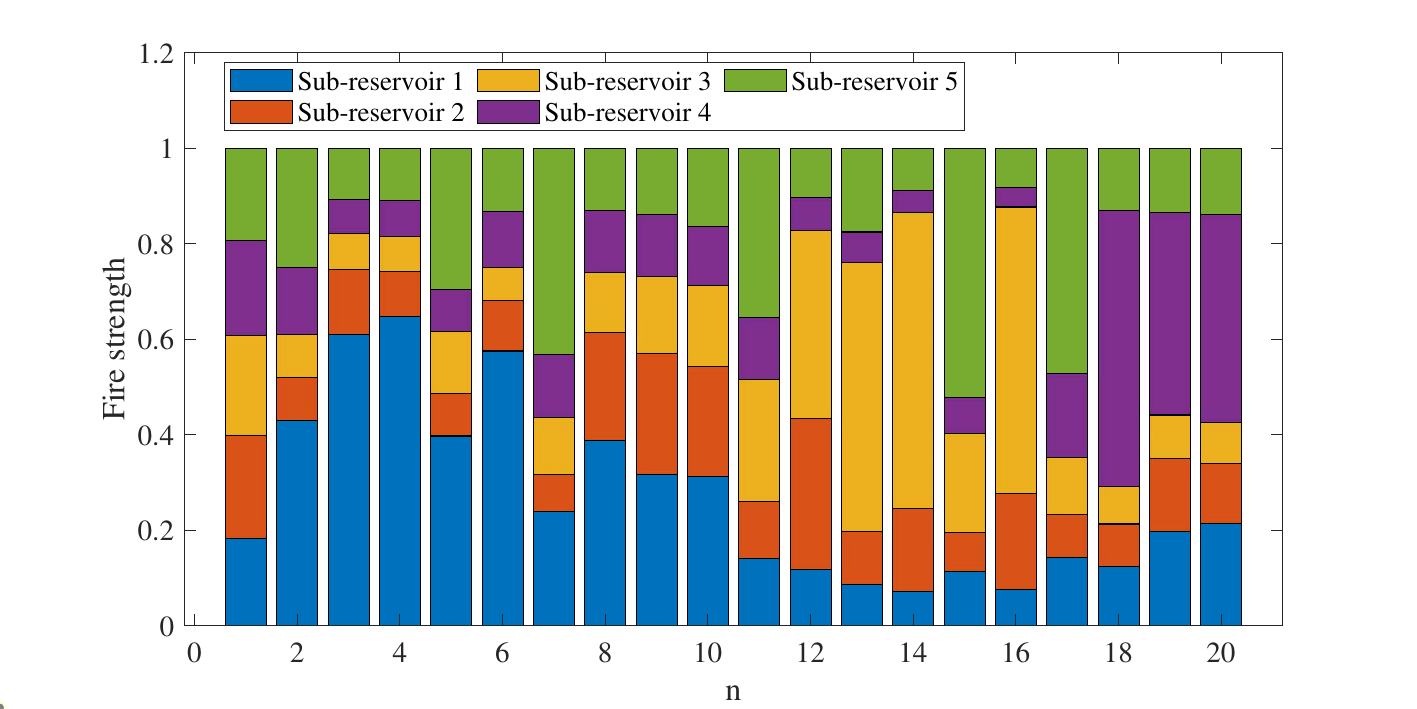}}
	\subfloat[Case 2]{\includegraphics[width=3.5in]{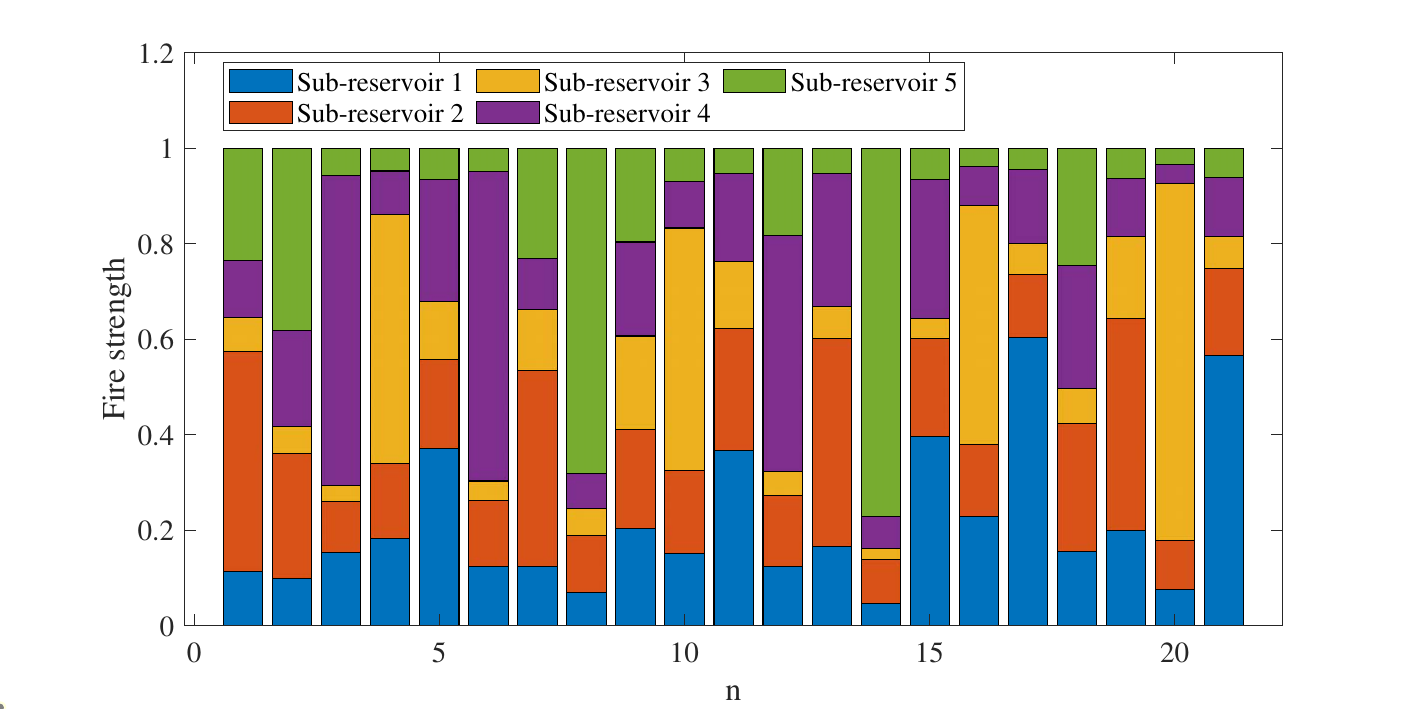}}
\caption{The fire strength of the F-RSCN for different testing samples on the two industry cases (\textit{Q}=5).}
\label{fig10}
\end{figure*}

Fig.~\ref{fig10} illustrates the fire strength of the F-RSCN for various testing samples in two industry cases, with sampling frequencies of 20 and 40 respectively. In case 1, the fire strength of sub-reservoir 1 initially starts high and gradually decreases, indicating strong predictive ability at the beginning that diminishes over time or with changing data. In case 2, the fire strength of each sub-reservoir fluctuates, suggesting varying adaptability or contribution at different time steps. The F-RSCN operates as an integrated model, where fire strength reflects the contribution of each sub-model. By incorporating feature information of input data through fuzzy logic reasoning, the model can identify more useful sub-reservoirs, influencing their selection and weight allocation. The integration of prior knowledge enhances the model's decision-making consistency with domain experts' understanding and expectations, thereby increasing credibility and interpretability. This makes the proposed F-RSCNs particularly effective for handling complex industrial process data.

\subsection{Comparisons and discussions}
Based on the theoretical analysis and simulation results reported above, this subsection briefly explains why our proposed F-RSCNs outperform other models.
\begin{itemize}
  \item  
  The comparisons between F-RSCNs and RSCNs: RSCNs have the advantages of fast learning speed, easy implementation, and strong approximation for nonlinear maps. Unfortunately, RSCNs struggle to provide intuitive explanations for complex fuzzy systems, making them less ideal in certain scenarios requiring model interpretability. F-RSCNs incorporate the advantages of RSCNs in terms of fast learning and universal approximation performance, as well as the advantages of TSK fuzzy inference systems in processing fuzzy information, thus further enhancing the nonlinear approximation ability and interpretability of the model.   
  \item 
  The comparisons between F-RSCNs and F-SCNs: Both models exhibit fuzzy reasoning capabilities. F-SCNs operate under the assumption that the input order is predetermined, leading to challenges in modelling nonlinear systems with unknown dynamic orders. In contrast, F-RSCNs employ the reservoir to store the historical information and construct a special structure of the random feedback matrix, which hold the echo state properties, effectively addressing the issue of order uncertainty. 
  \item 
  The comparisons between F-RSCNs and F-ESNs: Thses models are reservoir computing frameworks, which can effectively tackle the problems of slow convergence and easy falling into local minima in BP-based methods and uncertain dynamic orders. Furthermore, they are neuro-fuzzy models, making them suitable for fuzzy rule-based tasks. However, F-ESNs suffer from the sensitivity of learning parameters and arbitrary structure. F-RSCNs use a data-dependent weight scaling factor $\lambda $ that keeps varying during the learning process, enabling the built model to handle multi-scale random basis functions. Moreover, the sub-reservoirs of F-RSCNs are incrementally constructed in the light of a supervisory mechanism, ensuring the model's universal approximation property and leading to a more compact reservoir topology. Experimental results demonstrate that the proposed F-RSCNs can achieve sound performance in terms of training and testing.
\end{itemize}

\section{Conclusion}
This paper introduces a novel neuro-fuzzy model termed F-RSCN for industrial data analytics, with theoretical guarantees on the convergence of learning parameters. The F-RSCN integrates the TSK fuzzy inference system with the RSCN, which is a hybrid framework with multiple sub-reservoirs. From the implementation perspective, first, the fuzzy c-mean clustering algorithm is applied to generate the fuzzy rules based on the given data. Then, an RSC-based enhancement layer is used to process the dynamic nonlinear information, where the random parameters and network structure are determined in the light of a supervisory mechanism. Through this construction, each TSK fuzzy rule is represented by a sub-reservoir, and the model output is a weighted sum of outputs from these sub-reservoirs. Only the output weights are updated by the projection algorithm to handle the unknown dynamic data, without training the fuzzy inference system. The feasibility and effectiveness of the proposed F-RSCNs are validated on the one nonlinear system identification task and two industrial applications. The experimental results demonstrate that compared with other models, F-RSCNs can achieve sound performance for both learning and generalization. 

It is worth noting that the simple online parameter tuning may not be sufficient to enhance the model's learning capability for nonstationary data. Therefore, exploring a self-organizing version of F-RSCNs could be beneficial in improving the online self-learning ability of the model when dealing with nonstationary data streams in future research \cite{ref35}. Another extension of the proposed F-RSCN framework is to develop a recurrent version of stochastic configuration machines (SCMs) \cite{ref36}, where we will stress the technical challenges in terms of lightweight model implementation, result interpretability and efficiently uncertain data modelling.

\end{CJK}
\end{document}